\documentclass[journal]{IEEEtran} 
\usepackage{changes}
\usepackage{graphicx}
\usepackage{subfigure}
\usepackage{multirow}
\usepackage{longtable}
\usepackage{algorithm}
\usepackage{algorithmic}
\usepackage{booktabs}
\usepackage{stfloats}
\usepackage{caption2}
\usepackage{amsthm,amssymb,amsmath}
\usepackage{epstopdf}
\usepackage{color}
\usepackage{rotating}
\usepackage{multirow}
\usepackage{threeparttable}
\usepackage{mathrsfs}
\usepackage{bbding}
\usepackage{pifont}
\usepackage{makecell}
\usepackage[colorlinks,
 linkcolor=blue,
 anchorcolor=blue,
 citecolor=blue]{hyperref}

\ifCLASSINFOpdf
\else
\fi

\begin{document}
	\title{Lightweight RGB-D Salient Object Detection from a Speed-Accuracy Tradeoff Perspective}

	\author{
			Songsong Duan, \IEEEmembership{Student Member,~IEEE,}
			Xi Yang, \IEEEmembership{Senior Member,~IEEE,}
			Nannan Wang, \IEEEmembership{Senior Member,~IEEE,}
			and Xinbo Gao, \IEEEmembership{Fellow,~IEEE}
		\thanks{
		This work was supported in part by the National Natural Science Foundation of China under Grants 62372348 and 62036007, in part by the Key Research and Development Program of Shaanxi under Grant 2024GXZDCYL-02-10, in part by Shaanxi Outstanding Youth Science Fund Project under Grant 2023-JC-JQ-53, in part by the  Shaanxi Province Core Technology Research and Development Project under Grant 2024QY2-GJHX-11, in part by the Fundamental Research Funds for the Central Universities under Grant QTZX23042. (Corresponding author: Xi Yang)
		
		Songsong Duan, Xi Yang, and Nannan Wang are with the State Key Laboratory of Integrated Services Networks, School of Telecommunications Engineering, Xidian University, Xi’an 710071, China (e-mail: duanss@stu.xidian.edu.cn; yangx@xidian.edu.cn; nnwang@xidian.edu.cn).
		
		Xinbo Gao is with the Chongqing Key Laboratory of Image Cognition, Chongqing University of Posts and Telecommunications, Chongqing 400065, China (e-mail: xbgao@mail.xidian.edu.cn).		
		}
	}



\maketitle
\begin{abstract}
   Current RGB-D methods usually leverage large-scale backbones to improve accuracy but sacrifice efficiency. Meanwhile, several existing lightweight methods are difficult to achieve high-precision performance. To balance the efficiency and performance, we propose a Speed-Accuracy Tradeoff Network (SATNet) for Lightweight RGB-D SOD from three fundamental perspectives: depth quality, modality fusion, and feature representation. Concerning depth quality, we introduce the Depth Anything Model to generate high-quality depth maps,which effectively alleviates the multi-modal gaps in the current datasets. For modality fusion, we propose a Decoupled Attention Module (DAM) to explore the consistency within and between modalities. Here, the multi-modal features are decoupled into dual-view feature vectors to project discriminable information of feature maps. For feature representation, we develop a Dual Information Representation Module (DIRM) with a bi-directional inverted framework to enlarge the limited feature space generated by the lightweight backbones. DIRM models texture features and saliency features to enrich feature space, and employ two-way prediction heads to optimal its parameters through a bi-directional backpropagation. Finally, we design a Dual Feature Aggregation Module (DFAM) in the decoder to aggregate texture and saliency features. Extensive experiments on five public RGB-D SOD datasets indicate that the proposed SATNet excels state-of-the-art (SOTA) CNN-based heavyweight models and achieves a lightweight framework with 5.2 M parameters and 415 FPS. The code is available at \textit{https://github.com/duan-song/SATNet}.
    

\end{abstract}
\begin{IEEEkeywords}
RGB-D Salient Object Detection, Depth Anything Model, Lightweight, Decoupled Attention, Dual Information Representation
\end{IEEEkeywords}
\IEEEpeerreviewmaketitle
\section{\textbf{INTRODUCTION}}
\IEEEPARstart{S}{alient} Object Detection (SOD) aim to identify the most distinctive targets in a scene, referring to human ability to locate objects of interest, and segmenting them completely. Since SOD methods locates areas of interest, it has been widely transferred to many tasks involving target localization and search, such as, such as image retrieval \cite{ref-6}, visual tracking \cite{ref-7}, medical image segmentation \cite{ref-8}, and semantic segmentation \cite{ref-9}. To address challenging scenes characterized by low texture contrast or terrible backgrounds, depth maps have been widely adopted to enhance the recognition and localization ability of SOD methods. With the proliferation of depth acquisition devices and the development of depth prediction methods, depth maps have become more conveniently accessible, which in turn has further propelled the progress of RGB-D SOD.

\begin{figure}
	\includegraphics[width=0.48\textwidth,height=9cm]{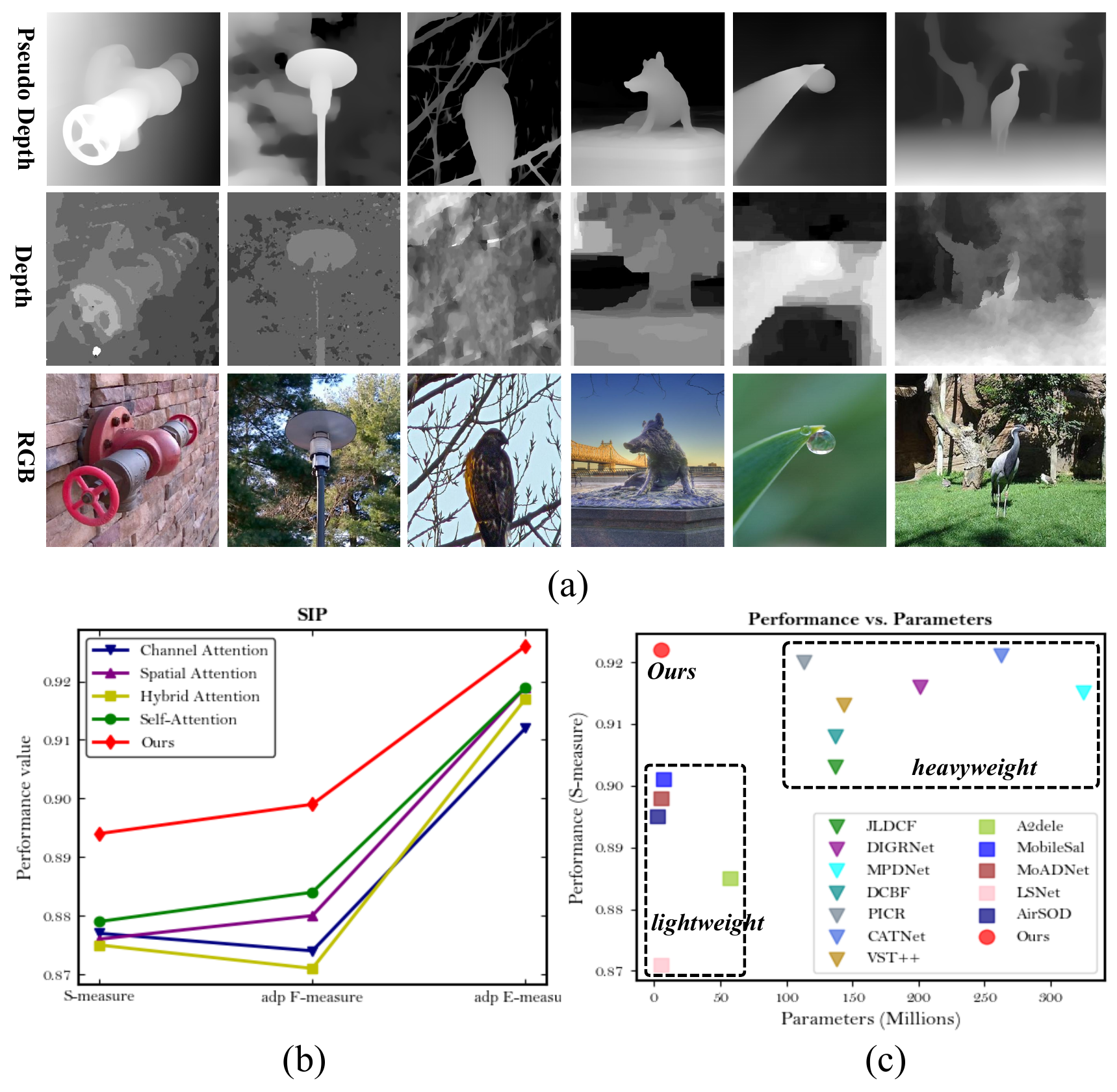}
	\caption{ Illustration of the motivation for our SATNet. (a) is the comparison of original depth maps and pseudo depth maps generated by the Depth Anything Model \cite{ref-81}; (b) is the different attention comparison; (c) is the comparison of performance and parameter of different RGB-D SOD methods.}
	\label{Fig.1}		
\end{figure}

 While significant progress has been made in accuracy metrics, the majority of current RGB-D SOD algorithms rely on high computational costs and large model sizes to achieve high performance. This approach severely hampers the deployment and real-time applications of RGB-D SOD algorithms, particularly on resource-constrained devices like smartphones and autonomous vehicles. To overcome these challenges, several lightweight RGB-D SOD approaches have been raised to develop efficient frameworks. However, the performance of these lightweight methods is typically subpar and fails to apply the practical scenarios. Therefore, this paper aims to explore RGB-D SOD methods to maximize performance within a lightweight paradigm to meet the demands of real-world applications. Specifically, we summarize the existing challenges of current lightweight methods and propose solutions to balance performance and efficiency from three vital perspectives:
 
\textbf{Low-quality Depth Maps.} Current lightweight RGB-D SOD approaches primarily focus on optimizing network structures to refine predicted saliency maps. However,the data quality in the training datasets has been overlooked. The low-quality of depth maps lead to the inconsistency between RGB and depth images existing in current datasets, which significantly impairs the performance of these methods. We present visualizations of depth maps from existing datasets in Fig. 1 (a), revealing the non-smooth depth values and noise points in poor depth maps.

 \textbf{Inadequate Modality Fusion.} Multi-modality fusion is a critical stage in the RGB-D SOD method. The heavyweight methods employ diverse attention mechanisms, such as channel \cite{ref-24}, spatial \cite{ref-25}, channel-spatial \cite{ref-25}, and self-attention \cite{ref-26}, to facilitate the fusion of RGB and depth information. Can the attention mechanisms utilized in heavyweight approaches be transferred to lightweight approaches and achieve success? We explore this issue and conduct a comparative analysis by comparing our design with these attention mechanisms in Fig. 1 (b). Obviously, these attention mechanisms cause an inadequate modality fusion in a lightweight framework, resulting in low performance.

 \textbf{Constrained Feature Representation.} Lightweight backbone networks hinder feature representation and inference performance. In contrast, heavyweight methods achieve superior performance with more parameters and channels. We think that the superiority of heavyweight methods is affluent feature space, such as parameters and channels. Specifically, the relationship between feature space and performance is shown in Fig. 1 (c), which reflects a positive correlation. Thus, the question arises: how to enhance the limited feature space in lightweight networks?
 
 To address the aforementioned issues, we rethink the development of lightweight RGB-D SOD and propose a Speed-Accuracy Tradeoff Network (SATNet) from the perspectives of depth quality, modality fusion, and feature representation. Firstly, we introduce the Depth Anything Model \cite{ref-81}, a vision foundation model for depth estimation with robust zero-shot generalization capabilities, into the RGB-D SOD to generate reliable depth maps. Secondly, to effectively integrate multi-modality features, we propose a Decoupled Attention Module (DAM) in a lightweight framework, which decouples modality-specific features into dual-view vectors to extract discriminative cues and then employs a cross interaction approach to enhance cross-modal features. DAM efficiently transforms 2D features into 1D vectors to accommodate lightweight models. Thirdly, we introduce a Dual Information Representation Module (DIRM) to extract texture and saliency features with two-way prediction heads. These features effectively enrich the constrained feature space, and the two-way prediction heads facilitate bi-directional parameter optimization. Finally, the Dual Feature Aggregation Module (DFAM) is developed to integrate texture and saliency features and furtherly reasoning saliency maps. Experiments on five public RGB-D SOD datasets verify the superiority of SATNet. Furthermore, Our SATNet gets more accurate saliency maps than other lightweight approaches by a wide margin. The key contributions of this work is summarized as follows:
\begin{itemize}
\item[$\bullet$] We develop a Speed-Accuracy Tradeoff Network, named \textit{SATNet}, which achieves more accurate performance compared with lightweight methods with much fewer model parameters (5.2 M) and faster inference speed (415 FPS).
	
\item[$\bullet$] We revisit the quality of depth maps in the current RGB-D SOD datasets, and introduce the Depth Anything Model to generate credible depth maps for addressing this issue.

\item[$\bullet$] We emphasize the importance of transitioning from heavyweight to lightweight models for attention mechanism, and propose a lightweight Decoupled Attention Module (DAM) to fuse RGB and depth features.

\item[$\bullet$] We propose a Dual Information Representation Module (DIRM), which enlarges constrained feature space via extracting texture and saliency features and achieves bi-directional optimization via two-way prediction heads.
\end{itemize}

The main purpose of introducing depth images into the SOD community is to provide clear, effective, and discriminative prior knowledge to assist SOD in complex scenarios. However, existing depth maps do not provide this prior knowledge, which does not meet the original intention of introducing depth maps. Using the depth-anything model to generate high-confidence depth maps can be justified. The reasons are as follows: (1) In many real-world scenarios, the actual captured depth map may suffer from noise, inaccuracies, or missing data due to limitations in depth sensors. The pseudo depth maps generated by a Depth Anything Model, trained on large and diverse datasets, can provide more consistent and reliable depth maps, especially in challenging environments. (2) The RGB-D SOD task mainly uses the texture and geometric prior knowledge contained in depth maps to process complex scenes, rather than the depth prior required by the 3D understanding task contained in the depth map. As a result, the pseudo depth map is more advantageous for SOD, which is a 2D segmentation task, and its practicality and usage scenarios are not easily limited.

 \section{Related Work}
 
 \subsection{\textbf{Salient Object Detection}}
 SOD has been fully studied for many years, yielding significant advancements. Traditional SOD methods primarily rely on hand-crafted features. However, because of restricted representation ability of hand-crafted features, traditional methods struggle with challenging scenes. In recent years, Convolutional Neural Network (CNN) based SOD models \cite{ref-37, ref-38} have gained prominence due to their superior feature extraction capabilities. Moreover, Transformer models have garnered attention for their exceptional global modeling abilities. Several Transformer-based SOD methods \cite{ref-34, ref-35, ref-41} are gradually proposed and achieve better performance than CNN-based ones. In this article, we focus on lightweight RGB SOD models. For example, PoolNet \cite{ref-1} emphasized pooling layers in U-shape architectures to refine high-level semantic features without additional parameters. Wang \textit{et al.} \cite{ref-91} proposed a context gating module (CGM) to learn visual attention mechanism of human. Li \textit{et al.} \cite{ref-2} introduced a depth-wise non-local module, which integrates lightweight convolution block and non-local module. 
 
 \subsection{\textbf{RGB-D Salient Object Detection}}
 Different from RGB SOD, RGB-D SOD focuses more on depth information extraction and collaborative learning. Thanks to the vast advance of CNN and Transformer networks in deep learning, many heavyweight RGB-D SOD methods, \textit{e.g.}, CNN-based and Transformer-based methods, achieve high accuracy. Zhang \textit{et al.}  \cite{ref-76} thought that current imaging technology to capture RGB and depth images results in inconsistent visual foreground layout. Therefore, they designed a depth feature calibration strategy to overcome the poor quality issue of depth maps by fusing original and pseudo depth maps. Chen \textit{et al.} \cite{ref-92} designed a multi-modal complementarity decoupled framework to decrease cross-modal ambiguity, which contains two specific decoupled elements: context disentanglement and representation disentanglement. Chen \textit{et al.} \cite{ref-117} introduced a 3-D convolution neural network to effectively promote the full integration of RGB and depth stream. Li \textit{et al.} \cite{ref-93} developed a scribble-labeling RGB-D saliency detection method to eliminate the dependence on accurate pixel-level annotations, which devised an online high-quality pseudo saliency mask generation mechanism to enhance the limited supervision information. To focus on depth information, some researchers designed depth-aware asymmetrical frameworks to discriminatively deal with RGB and depth images, aiming to excavate effective depth cues and explore consistency between multi-modality features. Cong \textit{et al.} \cite{ref-84} proposed a efficient cross-modal interaction module to extract depth feature and enhance RGB feature. To this end, Cong \textit{et al.} leveraged the local invariance of CNN and global dependencies of Transformer to build a cross-model point-aware interaction module. Wu \textit{et al.} \cite{ref-37} focused on the fine-gained semantic information of depth maps and developed a hierarchical depth-aware network (HiDAnet), which decomposed the depth map into different semantic masks by presetting different depth thresholds. Chen \textit{et al.} \cite{ref-96} proposed a modality transfer strategy to learn and extract the semantic information within and between modalities, excavating the complementarity priors via point-to-point matching. Luo \textit{et al.} \cite{ref-95} designed a hierarchy multi-scale feature aggregation method to fuse RGB and depth features. In addition, Piao \textit{et al.} \cite{ref-50} adopted the idea of knowledge distillation to distill the salient cues of depth maps into the RGB branch, thus realizing a lightweight detection framework. Besides, Mou \textit{et al.} \cite{ref-118} jointed RGB-depth and video data to achieve RGB-D Video SOD task and build a new dataset, name RDVS.

 \subsection{\textbf{Lightweight RGB-D Salient Object Detection}}
 Most of the traditional RGB-D SOD approaches employed large-scale convolutional neural networks and vision transformers to extract deep multi-modality features, like ResNet101, ConNeXt, Inception, Swin Transformer, and PVT. However, these neural networks are typically challenging to deploy on edge devices with limited resources and computing power, greatly hindering the practical application of RGB-D SOD methods. Fortunately, several lightweight and efficient neural networks were proposed and achieved real-time inference, like EfficientNet, MobileNet, and ShuffleNet. Benefiting from these networks, the researcher gradually proposed several lightweight RGB-D SOD methods. The standard dual-branch framework is a classical architecture for multi-modal SOD. Zhou \textit{et al.} \cite{ref-51} proposed a lightweight spatial boosting network (LSNet) with a boundary boosting algorithm to enhance edge details of predicated saliency maps. AirSOD \cite{ref-16} designed a hybrid feature extraction network to explore RGB and depth features, which contains a lightweight MobileNet V2 and a parallel attention-shift (PAS) convolution module. Another approach involves employing a depth-tailored strategy to learn elaborate saliency features. For instance, A2dele \cite{ref-50} designed a two-stage saliency feature learning strategy, which first extracted salient attention maps from depth maps and then refined the RGB feature via salient attention maps. Wu \textit{et al.} \cite{ref-52} found that most lightweight backbones (like MobileNet) lacked powerful feature representation, which led to an uncompetitive performance. Therefore, they focused on the potential of depth maps and proposed an implicit representation learning technology. Jin \textit{et al.} \cite{ref-15} leveraged contextual receptive field information to refine fine-grained features of mobile neural networks. Specifically, they attempted to set different dilated convolutions on hierarchical blocks and aggregated these dilated features with different contexts. Zeng \textit{et al.} \cite{ref-16} proposed a mobile asymmetric dual-stream network to fuse multi-modality features with an inverted bottleneck structure.

 However, there is still a big gap between these methods and the heavyweight algorithms. To reduce the performance gap with heavyweight methods, we rethink the design of lightweight RGB-D SOD methods and improve lightweight model via depth quality, multi-modality fusion, and feature representation.

  \begin{figure*}
  	\centering\includegraphics[width=0.85\textwidth,height=8cm]{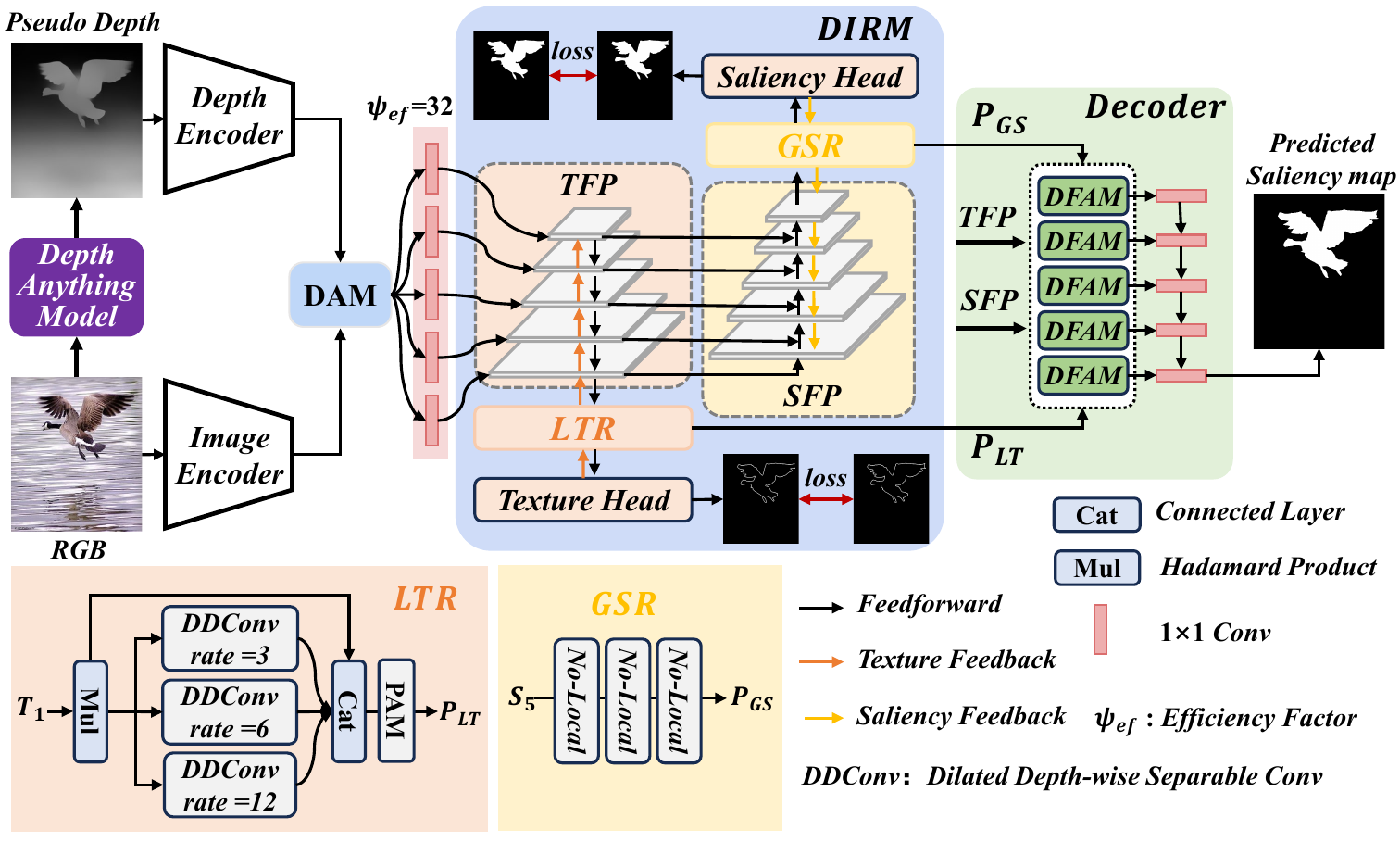}
  	\caption{Framework of proposed SATNet. It consists of four parts, including RGB and Depth Encoders, Decoupled Attention Modules (DAM), a Dual Information Representation Module (DIRM), and a Decoder with the Dual Feature Aggregation Modules (DFAM). Note that we employ the depth anything model to generate pseudo depth map with more high quality than original one.}
  	\label{Fig.2}
  	\vspace{-5mm}		
  \end{figure*}

\section{\textbf{Methodology}}

\subsection{\textbf{Overview}}
We show an overview of the proposed SATNet in Fig. 2, which consists of four components, including RGB and Depth Encoders, Decoupled Attention Modules (DAM), a Dual Information Representation Module (DIRM), and a decoder with Dual Feature Aggregation Module (DFAM). 

First, we discuss the quality issue of depth maps existing in current RGB-D SOD datasets and introduce the Depth Anything Model to generate more reliable depth maps. The existing lightweight methods ignore the quality of depth maps, resulting in unsatisfactory performance compared with heavyweight methods. To address this issue, we introduce the Depth Anything Model \cite{ref-81}, a vision foundation model for monocular depth estimation, to produce credible pseudo depth maps. Fig. 1 (a) exhibits some visual comparison of depth maps in existing datasets and pseudo depth maps generated by the Depth Anything Model. Depth Anything Model is trained in a large-scale dataset, containing 62 million images collected from other public large-scale datasets, \textit{e.g.}, SA-1B, OpenImages, and BDD100K. The large-scale training data causes the powerful zero-shot generalization capabilities for the Depth Anything Model, which guarantees the reliability of pseudo depth maps.

After processing with the Depth Anything Model, we obtain higher quality depth maps than original ones. Then, we send the RGB and corresponding pseudo depth maps into RGB and depth encoders to extract RGB and depth features, namely $\{ f_{r}^{i}\}_{i=1}^{5}$ and $\{ f_{d}^{i}\}_{i=1}^{5}$. Specifically, RGB and depth encoders employ MobileNet v2 \cite{ref-21} to achieve a lightweight framework. Then, we adopt a $1\times 1$ convolution operation to unify the channels of $\{ f_{r}^{i}\}_{i=1}^{5}$ and $\{ f_{d}^{i}\}_{i=1}^{5}$ for reducing the complexity of the subsequent processing, namely the efficiency factor $\psi_{ef}=32$. Next, we use DAM to fuse RGB and depth features, named as $f_{f}^{i}\in\mathbb{R}^{C \times H \times W}$. Then, DIRM builds a efficient and effective dual feature learning unit to model texture and saliency features, called $ \{T_{i}\}_{i=1}^{5}$ and $ \{S_{i}\}_{i=1}^{5} \in \mathbb{R}^{ef \times \frac{H}{2^i} \times \frac{W}{2^i}}$, respectively. Finally, the $\{T_{i}\}_{i=1}^{5}$ and $\{S_{i}\}_{i=1}^{5}$ are sent into DFAM of decoder to integrate texture and saliency features and reasoning saliency maps.

\subsection{\textbf{Decoupled Attention Module}}

The depth maps reveal spatial information of RGB images, which distinguishes the targets from the background, especially for challenging scenes. Existing fusion strategies employ general attention to guide the integration of RGB and depth features, which neglects insufficient feature space and feature representation abilities. Unlike these models, we propose a lightweight attention mechanism for lightweight models to integrate cross-modal features.

As shown in Fig. 3, DAM considers the insufficient feature space and feature representation. Specifically, we decouple the input features into horizontal and vertical vectors through decoupled pooling to capture fine-grained pixels:
\begin{equation}
\left\{
\begin{array}{lr}
\mathcal{R}_{h}^{i} = \mathcal{DP}_{h}(f_{r}^{i}),  \\

\mathcal{R}_{v}^{i} = \mathcal{DP}_{v}(f_{r}^{i}),
\end{array}
\right.
\end{equation}
where $\mathcal{R}_{h}^{i} \in\mathbb{R}^{C \times 1 \times W} $ and $\mathcal{R}_{v}^{i} \in\mathbb{R}^{C \times H \times 1} $ are the horizontal and vertical vectors of RGB features, respectively. Similarly, $\mathcal{D}_{h}^{i} \in\mathbb{R}^{C \times 1 \times W} $ and $\mathcal{D}_{v}^{i} \in\mathbb{R}^{C \times H \times 1} $ are the horizontal and vertical vectors of depth features. $\mathcal{DP}_{h}(.)$ and $\mathcal{DP}_{v}(.)$ indicate the adaptive pooling operations along with the width and height of feature maps for decoupling original features. $C$ means the channel numbers of features. After that, we connect the two vectors and then employ linear and nonlinear transformation to project the attention weights of the dual-view features:
\begin{equation}
\mathcal{\hat{R}}_{h}^{i}, \mathcal{\hat{R}}_{v}^{i} = \Pi_{2}\{\mathcal{NL}_{6}(\mathcal{FC}^{2}(\mathcal{BN}(\mathcal{C}at(\mathcal{R}_{h}^{i},\mathcal{R}_{v}^{i}))))\},
\end{equation}
\begin{equation}
	\mathcal{\widetilde{R}}_{h}^{i}, \mathcal{\widetilde{R}}_{v}^{i} = \mathcal{SIG}(\mathcal{\hat{R}}_{h}^{i}), \mathcal{SIG}(\mathcal{\hat{R}}_{v}^{i}),
\end{equation}
\begin{equation}
V_{r}^{i} = \mathcal{MUL}_{3}\{f_{r}^{i}|(\mathcal{\widetilde{R}}_{h}^{i}, \mathcal{\widetilde{R}}_{v}^{i})\},
\end{equation}
where $\mathcal{MUL}_{3}()$ denotes the hadamard product with three input elements. $\mathcal{CAT}()$ is the connected layer for connecting $\mathcal{R}_{h}^{i}$ and $\mathcal{R}_{v}^{i}$ via a permuting transformation, where the shape of $\mathcal{R}_{v}^{i}$ is transmitted from $\mathbb{R}^{C \times H \times 1}$ to $\mathbb{R}^{C \times 1 \times H}$. Hence, the shape of connected vectors is $\mathbb{R}^{C \times 1 \times (W + H)}$. $\mathcal{FC}^{2}$ indicates two fully connected layers to learn the importance of each width and height coordinate in the feature maps. $\mathcal{BN}$ is the batch normalization. $\mathcal{NL}_{6}$ denotes the ReLU6 activation function, which is a nonlinear operation for reinforcing sparse sampling. $\Pi_{2}\{\}$ reflects the split function with an opposite permute layer. $\mathcal{SIG}$ denotes the sigmoid function. $\mathcal{\widetilde{R}}_{h}^{i}$ and $\mathcal{\widetilde{R}}_{v}^{i}$ are dual-view attention vectors for enhancing multi-modality features. Then, the two attention vectors are employed to filter out original input features $f_{r}^{i}$ to generate the refined RGB features $V_{r}^{i}$. Similarly, we compute the refined depth features $V_{d}^{i}$ with the same operations.
\begin{figure}
	\centering\includegraphics[width=0.49\textwidth,height=5.5cm]{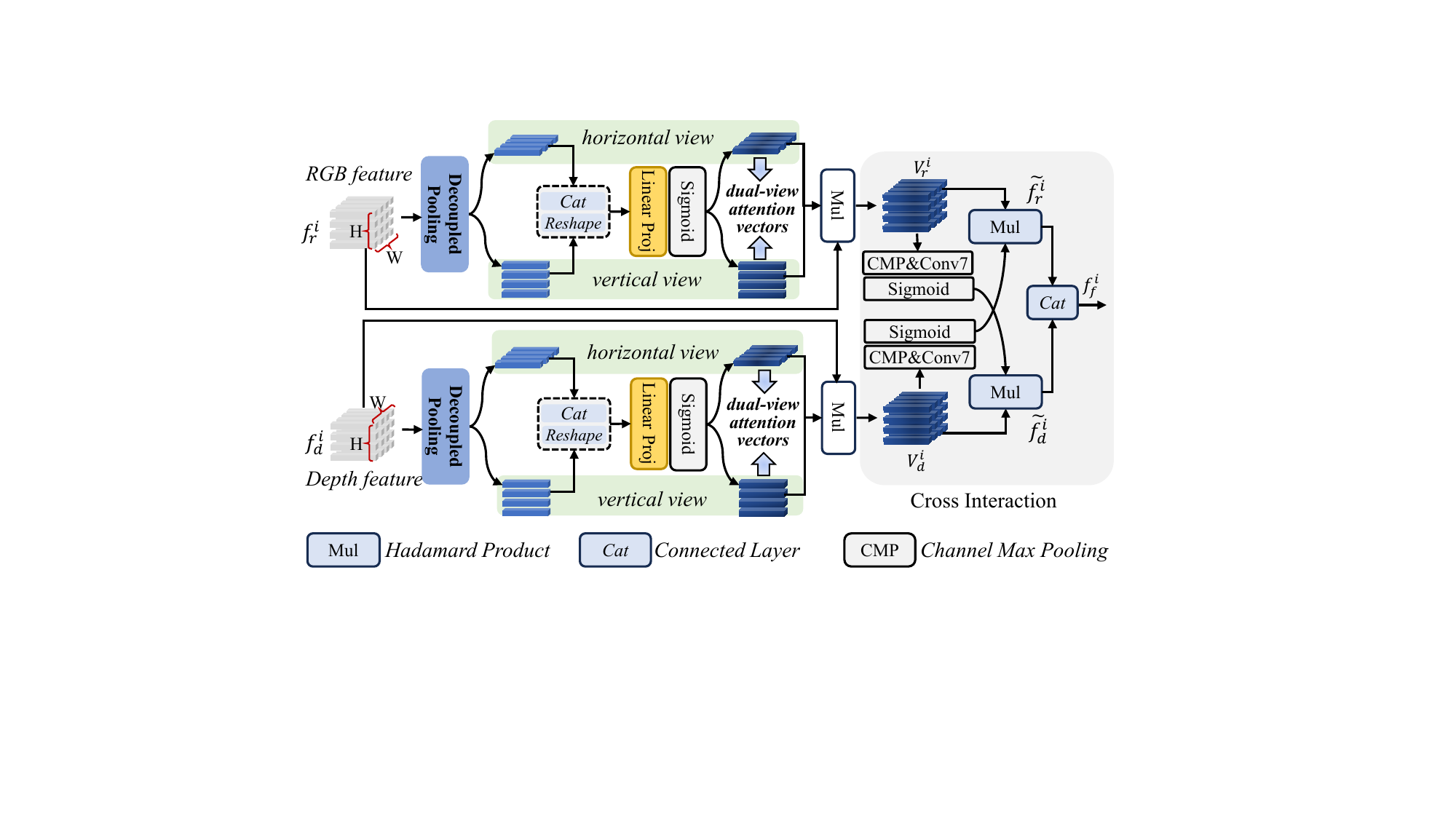}
	\caption{The illustration of the proposed DAM. The CMP is max pooling along with channels.}
	\label{Fig.3}		
\end{figure}
\begin{figure}
	\centering\includegraphics[width=0.45\textwidth,height=4.5cm]{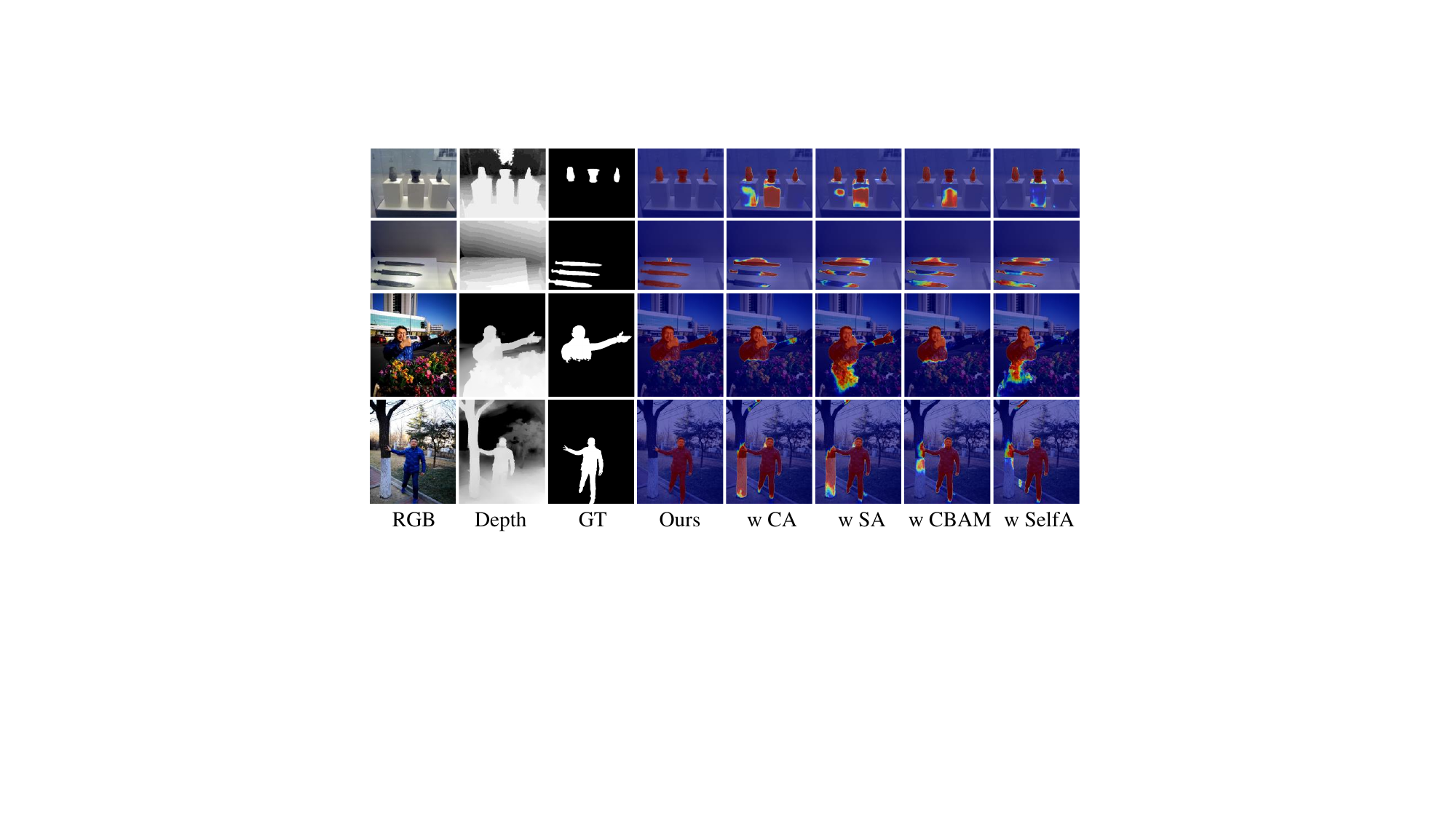}
	\caption{The visualization of the proposed DAM and other variants with channel Attention (CA), Spatial Attention (SA), Hybrid Attention (CBAM), and Self-Attention (SelfA).}
	\label{Fig.4}		
\end{figure}

Cross-modal interaction encourages consistency learning between RGB and depth modalities. Following this principle, we employ a cross interaction strategy to integrate refined RGB and depth features. Specifically, we squeeze the refined RGB features $V_{r}^{i}$ for global information embedding, which learns the significant descriptor by a max pooling along with channel:
\begin{equation}
W_{r}^{i} = \mathcal{SIG}(\mathcal{C}onv_{7}(\mathcal{POOL}_{max}(V_{r}^{i}))),
\end{equation}
where $\mathcal{POOL}_{max}()$ is the max pooling along with the channel. $\mathcal{C}onv_{7}$ denotes a convolutional layer with $7 \times 7$ kernel to enlarge the receptive field. $W_{r}^{i} \in \mathbb{R}^{1\times \frac{H}{2^i} \times \frac{W}{2^i}}$ serves as a spatial heatmap to highlight  discriminative regions within feature maps. The depth branch repeats the same operations to generate the $W_{d}^{i} \in \mathbb{R}^{1\times \frac{H}{2^i} \times \frac{W}{2^i}}$. Then, cross multiplication are used to fuse RGB and depth features with the spatial heatmap:
\begin{equation}
	\left\{
	\begin{array}{lr}
		\widetilde{f}_{r}^{i} = \mathcal{MUL}_{2}(f_{r}^{i}|W_{d}^{i}),  \\
		\widetilde{f}_{d}^{i} = \mathcal{MUL}_{2}(f_{d}^{i}|W_{r}^{i}),
	\end{array}
	\right.
\end{equation}
where $\mathcal{MUL}_{2}()$ denotes the hadamard product with two input elements. $\widetilde{f}_{r}^{i}$ and $\widetilde{f}_{d}^{i}$ indicate the enhanced RGB and depth features through cross-modal interaction. Finally, we use channel max pooling operation to select maximized channel feature map between $\widetilde{f}_{r}^{i}$ and $\widetilde{f}_{d}^{i}$, and then generate the fused multi-modality features $f_{f}^{i}$.

\textbf{\textit{Visualization}}. Our DAM obtains more promising performance than other attention mechanisms in a lightweight setting. To further verify the effectiveness of our proposed DAM, we present the visual comparison in Fig. 4, including channel attention \cite{ref-24}, spatial attention \cite{ref-25}, hybrid attention \cite{ref-25}, and self-attention \cite{ref-26}. It can be seen that our model focus on more complete saliency regions compared with other attention mechanisms, which proves the effectiveness of our DAM.

\begin{figure}
	\centering\includegraphics[width=0.48\textwidth,height=5cm]{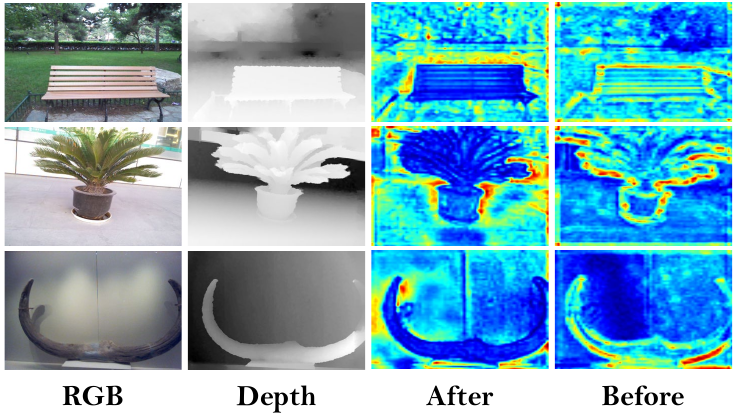}
	\caption{The feature visualization of the proposed DIRM, including the features before and after the DIRM.}
	\label{Fig.6}		
\end{figure}

\subsection{\textbf{Dual Information Representation Module}}

Representation learning is a fundamental technique in deep learning and computer vision tasks, like prompt learning, self-supervised learning, and semantic segmentation. Many works attempt to enhance the representation learning. For example, ASPP \cite{ref-90} and DenseASPP \cite{ref-89} leverage dilated convolutions to effectively capture receptive field information. Furthermore, the token representation of Vision Transformer (ViT) \cite{ref-53} improves the global long-range dependencies. In this article, we discuss the representation learning of lightweight networks and propose a Dual Information Representation Module (DIRM) to enrich the constrained feature space in a lightweight model by a dual-feature representation. DIRM contains a Semantic Feature Pyramid (SFP), a Texture Feature Pyramid (TFP), a Global Semantic Refinement (GSR), and a Local Texture Refinement (LTR). As shown in Fig. 2, this comprehensive architecture aims to enhance feature representation within the lightweight framework.

 \begin{figure}
	\centering\includegraphics[width=0.50\textwidth,height=6cm]{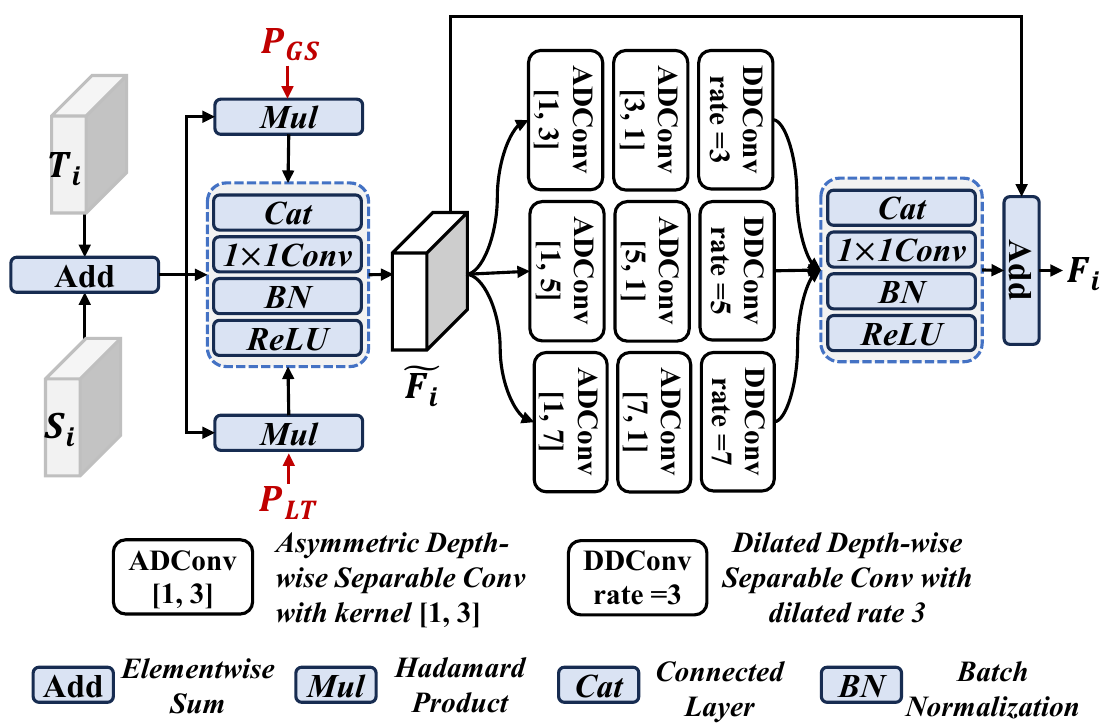}
	\caption{The illustration of the proposed DFAM.
	}
	\label{Fig.5}		
\end{figure}

\textit{\textbf{TFP and LTR}}. TFP extracts texture features in a top-down manner, which complements the global semantic deficiency of low-level features under the supervision of Edge Ground Truth (GT). Then, we utilize the LTR module to refine the bottom features of TFP, where the LTR module effectively learns the local texture prior $P_{LT}$. Specifically, we firstly integrate the multi-modality fused features $f_{f}^{i}$ via connected layer and $1 \times 1$ convolution operation. The total processing is defined as:
\begin{equation}
	\left\{
	\begin{array}{lr}
		T_{i} = f_{f}^{i}, \quad \quad \quad  \quad \quad \quad  \quad \quad \quad  \quad  \quad i=5, \\
		
		T_{i} = \mathcal{C}onv_{1}(\mathcal{CAT}(f_{f}^{i}, \mathcal{UP}(T_{i+1}))), i \in \{1,2,3,4\},
	\end{array}
	\right.
\end{equation}
where $\mathcal{UP}()$ indicates the $2 \times$ up-sampling operation; $\mathcal{C}onv_{1}$ is a $1 \times 1$ convolution operation. After the interaction of high-level features, we obtain a favorable texture feature pyramid via the supervision of Edge GT, as shown in Fig. 2. Then, we design an LTR module to extract the local texture prior $P_{LT}$. Given resolution in the lowest level feature, the dilated convolutions are employed to gain rich receptive field information via sparse sampling. Concretely, three parallel dilated convolution layers with different dilated rates are embedded into the LTR. Next, we leverage the off-the-shelf patch attention module (PAM) \cite{ref-5} to capture the refined local detail information, which is tailored for enhancing local features.

\textbf{\textit{SFP and GSR}}. Unlike TFP, SFP uses down-top flow to complement local texture of high-level features via down-sampling operation. The top features with the smallest feature resolution are input to the GRS module to extract global long-range dependencies via non-local modules, where the GRS generates global semantic prior $P_{GS}$. Concretely, we first progressively integrate texture pyramid $\{T_{i}\}_{i=1}^{5}$ via down-sampling operations, which is defined as:
\begin{equation}
	\left\{
	\begin{array}{lr}
		S_{i} = T_{i}, \quad \quad \quad  \quad \quad \quad  \quad \quad \quad \quad  \quad \, \, i=1, \\
		
		S_{i} = \mathcal{C}onv_{1}(\mathcal{CAT}(T_{i}, \mathcal{DW}(S_{i-1}))), i \in \{2,3,4,5 \},
	\end{array}
	\right.
\end{equation}
where $\mathcal{DW}()$ is the $2 \times$ down-sampling operation; $\{S_{i}\}_{i=1}^{5}$ is the semantic features. Further, we leverage the non-local module \cite{ref-59} to ameliorate global semantic information. The GSR module is built by three stacked non-local modules for extracting global long-range dependencies. Note that the input features are the top features of SFP compared with TFP. The global semantic prior $P_{GS}$ is generated under the supervision of saliency GT.

\textbf{\textit{Two-way Prediction Heads}}. As depicted in Fig. 2, we design two prediction heads, namely the Texture Head and the Saliency Head, to optimize the parameters of DIRM. Specifically, the Texture Head provides the gradient information to TFP under the guidance of Edge GT, then generates the texture features. Similarly, the Saliency Head prompts the learning of saliency features via using Saliency GT. In this way, DIRM represents two types of representation information and enhances the feature space. To demonstrate the effectiveness of DIRM, we show feature visualizations in Fig. 5, including the features before and after DIRM. We find that the features after using DIRM are better than before.

\subsection{\textbf{Dual Feature Aggregation Module}}

After TFP and SFP, DIRM obtains two kinds of feature representation with texture and saliency information. As shown in Fig. 6, DFAM integrate the two feature with texture $P_{LT}$ and saliency prior $P_{GS}$, which defined as:
\begin{equation}
		F_{i}^{GS} =\mathcal{MUL}_{2}((T_{i} + S_{i})|P_{GS}),
\end{equation}
\begin{equation}
	F_{i}^{LT} =\mathcal{MUL}_{2}((T_{i} + S_{i})|P_{LT}),
\end{equation}
\begin{equation}
	\widetilde{F}_{i} = \mathcal{C}onv_{1}(\mathcal{CAT}(F_{i}^{GS}, F_{i}^{LT}, (T_{i} + S_{i}))),
\end{equation}
where $F_{i}^{GS}$ and $F_{i}^{LT}$ are the fused features with multiple information under the guidance of global semantic prior $P_{GS}$ and local texture prior $P_{LT}$, respectively. $\widetilde{F}_{i}$ is the final fused features with multiple information. $\mathcal{C}onv_{1}$ is the $1 \times 1$ convolution, which is used to recover channels of features after the $\mathcal{CAT}()$ operation.

 The receptive field information is a vital context cue for neural networks, which improves the representation of feature maps via large convolutional kernels. In a word, the receptive field is an essential part of convolutional neural networks. For example, ConvNeXt \cite{ref-20} outperforms Swin Transformer \cite{ref-54} by leveraging large $7 \times 7$ convolution. However, a large convolutional kernel must result in a larger parameter. Therefore, we adopt the asymmetric convolution \cite{ref-60} and dilated convolution \cite{ref-61} to implement our idea, which expands the receptive field without any additional parameters. 

We set up three branches to explore different receptive fields in DFAM, including $3 \times 3$, $5 \times 5$, and $7 \times 7$, which is formulated as:
\begin{equation}
	F_{i}^{Rk} = \mathcal{DDC}onv_{k}(\mathcal{ADC}onv_{1}^{k}(\mathcal{ADC}onv_{k}^{1}(\widetilde{F}_{i}))),
\end{equation}
where $k \in \{3,5,7\}$ indicates the different receptive fields; $\mathcal{ADC}onv_{k}^{1}$, $\mathcal{ADC}onv_{1}^{k}$, and $\mathcal{DDC}onv_{k}$ are $k \times 1$ asymmetric depth-wise separable convolution, $1 \times k$ asymmetric depth-wise separable convolution, and $3 \times 3$ dilated depth-wise separable convolution with dilated rate $k$, respectively. Finally, we fuse the three receptive field features $F_{i}^{R3}$, $F_{i}^{R5}$, and $F_{i}^{R7}$ to generate output features $F_{i}$ with multiple receptive fields.

\begin{table*}
	\large
	\renewcommand\arraystretch{2.5}
	\centering
	\fontsize{8}{8}
	\selectfont
	\caption{\label{comparison} Quantitative comparisons of the proposed method against the other 16 SOTA RGB-D SOD methods, including 10 high-complexity RGB-D SOD methods and 6 lightweight RGB-D SOD methods. $\uparrow$/$\downarrow$ indicates that a larger/smaller is better. The best results of lightweight RGB-D SOD methods are highlighted in \textbf{bold}.}
	\label{tab:distortion_type}
	\setlength{\tabcolsep}{2pt} 
	\renewcommand\arraystretch{1.1}
	\resizebox*{\textwidth}{!}{
		\large
		\begin{tabular}{cc|cccccccccc|ccccccc}
			\toprule[2pt]
			
			\multicolumn{2}{c|}{\textbf{Method }} &
			\multicolumn{1}{c}{\textbf{DSA2F }} &
			\multicolumn{1}{c}{\textbf{BiANet}} &
			\multicolumn{1}{c}{\textbf{JLDCF }} &
			\multicolumn{1}{c}{\textbf{DIGR  }} &

			\multicolumn{1}{c}{\textbf{MPDNet}} &
			\multicolumn{1}{c}{\textbf{HiDA  }} &
			\multicolumn{1}{c}{\textbf{HINet }} &
			\multicolumn{1}{c}{\textbf{DCBF }} &
			\multicolumn{1}{c}{\textbf{FCFNet}} &
			\multicolumn{1}{c|}{\textbf{DGFNet}} &
			
			\multicolumn{1}{c}{\textbf{A2dele }} &
			\multicolumn{1}{c}{\textbf{DFMNet }} &
			\multicolumn{1}{c}{\textbf{MobileSal}} &
			\multicolumn{1}{c}{\textbf{MoADN  }} &
			\multicolumn{1}{c}{\textbf{LSNet  }} &
			\multicolumn{1}{c}{\textbf{AirSOD }} &
			\multirow{3}{*}{\makecell[c]{\textbf{\textit{SATNet}} \\ \textbf{Ours}}}
			\cr
			
			\multicolumn{2}{c|}{\textbf{Pub. Year }}	
			& 2021 & 2021 &	2022 & 2022	&	2023 & 2023	&	2023 &	2023 & 2024 & 2024 	
			
			& 2020 & 2021 & 2022 & 2022 & 2023 & 2024 \cr
			
			\multicolumn{2}{c|}{\textbf{Publication }} 
			& CVPR & TIP & TPAMI & TMM & TCSVT	& TIP & PR & IJCV &	TCSVT	& TCSVT &	CVPR &	MM & TPAMI & TIP &	TIP & TCSVT \cr
			
			\hline    
			\multicolumn{2}{c|}{\textbf{Params(M)}}	
			& - & 50 &	137 & 201	&	325 & 130.6	&	99 & 137 & 42 & - & 57.3 & 8.5 & 6.5 & 5 & 5.4 & 2.4 & 5.2 \cr  
			\multicolumn{2}{c|}{\textbf{ FLOPs(G) }}	
			& - & 58.4 &	736.1 & -	&  52 & 71.5	&	- &	-	& 74.9 & - & 41.9 & 2.5 & 1.6 & 1.3 & 1.2 & 0.9 & 1.5  \cr  
			\multicolumn{2}{c|}{\textbf{Speed(FPS) }}	
			& - & 34 &	9 & 20	& 18 & 57	&	16.3 &	9	& - & - & 120 & 236 & 290 & 80 & 93.5 & 365 & 415  \cr        
			
			\hline
			\multicolumn{2}{c|}{\textbf{Type}} & \multicolumn{10}{|c|}{\textbf{Heavyweight RGB-D SOD Methods}} & \multicolumn{7}{c}{\textbf{Lightweight RGB-D SOD Methods}} \cr
			\hline
			
			\multirow{5}{*}{\textbf{\rotatebox{270}{NLPR}}}
			&\textbf{$adpEm\uparrow$}    
			& 0.952 & 0.939 & 0.955 & 0.957  &	0.951 &	0.961 &	0.950 &	0.956 &  - &	0.949  & 0.945 &	0.954 &	0.955 &	0.947 &	0.956 &	0.957 &	\textbf{0.964} \cr
			
			&\textbf{$Sm\uparrow$}       
			& 0.917 & 0.900 & 0.925 & 0.935  &	0.932 &	0.929 &	0.922 &	0.921 & 0.923 & 0.924 & 0.896 & 0.925 & 0.919 & 0.915 & 0.918 & 0.925 & \textbf{0.932} \cr
			
			&\textbf{$adpFm\uparrow$}  
			&0.896&	0.849 &	0.878 &	0.890  &	  -  & 0.908 &	0.877 &	0.893 &	0.893 &	0.879 & 0.878 & 0.880 & 0.878 &	0.875 &	0.885 &	0.884 &	\textbf{0.900}\cr
			
			&\textbf{$WF\uparrow$}  
			&0.889&	0.833 &	0.882 &	0.890  &	0.868 &	0.901 &	0.871 &	0.892 &	-   &  -  & 0.867 & 0.877 & 0.870 &	0.875 &	0.883 &	0.881 &	\textbf{0.897}\cr
			
			&\textbf{$MAE\downarrow$ }          
			&0.024&	0.032 &	0.022 &	0.023  &	0.025 &	0.021 &	0.026 &	0.023 &	0.025 &	0.024 & 0.028 & 0.024 & 0.024 &	0.027 &	0.024 &	0.023 &	\textbf{0.019}\cr

			\hline
			\multirow{5}{*}{\textbf{\rotatebox{270}{NJU2K}}}
			&\textbf{$adpEm\uparrow$}    
			& 0.937&	0.907 &	0.935 &	0.954 &   -  & 0.953 &	0.939 &	0.941 & - & 0.929 & 0.916 &	0.937 &	0.935 &	0.935 &	0.940 &	0.934 &	\textbf{0.952}  \cr
			
			&\textbf{$Sm\uparrow$}       
			& 0.903&	0.868 &	0.902 &	0.933  &	0.912  & 0.926 &    0.915 &	0.903 &	0.920 &	0.918	&0.869 &	0.912 &	0.896 &	0.906 &	0.911 &	0.908 &	\textbf{0.925} \cr
			
			&\textbf{$adpFm\uparrow$}  
			& 0.901&	0.848 &	0.885 &	0.917  &	 -  & 0.922 &	0.896 &	0.898 &	0.912 &	0.891 & 0.874 &	0.894 &	0.886 &	0.892 &	0.900 &	0.889 &	\textbf{0.917} \cr
			
			&\textbf{$WF\uparrow$}  
			& 0.889&	0.811 &	0.869 &	0.904 & 0.860  & 0.910 &	0.877 &	0.884 &  -  & - &0.851 &	0.879 &	0.861 &	0.881 &	0.887 &	0.879 &	\textbf{0.903} \cr
			
			&\textbf{$MAE\downarrow$ }          
			& 0.039&	0.056 &	0.041 &	0.028  & 0.041  & 0.029 & 0.039 &	0.038 & 0.033 &	0.034 & 0.051 &	0.039 &	0.044 &	0.041 &	0.038 &	0.039 &	\textbf{0.029}  \cr

			\hline
			\multirow{5}{*}{\textbf{\rotatebox{270}{SIP}}}
			&\textbf{$adpEm\uparrow$}    
			& 0.911&	0.866& 	0.923 &	0.919 &	0.925 &	0.927 &	0.899 &	0.920 &  -  & -   &	0.890& 	0.924 &	0.911 &	0.911 &	\textbf{0.928} &	0.904 &	0.926  \cr
			
			&\textbf{$Sm\uparrow$}       
			& 0.861&	0.803 &	0.880 &	0.885  &	0.900 &	0.892 &	0.856 &	0.873&	0.892 &	-  &    0.827& 	0.885 &	0.866 &	0.865 & 0.886 &	0.859 &	\textbf{0.894} \cr
			
			&\textbf{$adpFm\uparrow$}  
			& 0.867&	0.780 &	0.873 &	0.879 &	-  & 0.893 &	0.846 &	0.877 &	0.888 &-  &	0.829& 	0.874 &	0.859 &	0.850 &	0.884 &	0.855 &	\textbf{0.896} \cr
			
			&\textbf{$WF\uparrow$}  
			&  0.837&	0.718 &	0.844 &	0.841  &	0.854 &	0.866 &	0.796 &	0.849 &  -  & - &  0.794 &  0.842 &	0.820 &	0.828 &	0.857 &	0.814 &	\textbf{0.865} \cr
			
			&\textbf{$MAE\downarrow$ }          
			&  	0.056&	0.091 &	0.049 &	0.053 &	0.044 &	0.044 &	0.066 &	0.051 &	0.044 &	-   &	0.070& 	0.049 &	0.057 &	0.058 &	0.049 &	0.060 &	\textbf{0.044} \cr

			\hline
			\multirow{5}{*}{\textbf{\rotatebox{270}{STERE}}}
			&\textbf{$adpEm\uparrow$}    
			& 0.949&	0.925 &	0.937 &	0.943  &	0.936 &	0.945 &	0.927 &	0.945 &	 -	 &  0.927 &	0.935 &	0.939 &	0.938 &	0.935 &	0.919 &	0.932 &	\textbf{0.953}  \cr
			
			&\textbf{$Sm\uparrow$}       
			& 0.903&	0.888 &	0.903 &	0.916  &	0.915 &	0.911 &	0.892 &	0.908 &	0.885 &	0.908 &	0.913& 	0.906 &	0.901 &	0.898 &	0.871 &	0.895 &	\textbf{0.922} \cr
			
			&\textbf{$adpFm\uparrow$}  
			& 0.898&	0.869 &	0.869 &	0.889  & - &	0.897 &	0.859 &	0.897 &	0.894& 	0.868 & 0.884 &	0.875 &	0.878 &	0.868 &	0.854 &	0.865 &	\textbf{0.907} \cr
			
			&\textbf{$WF\uparrow$}  
			& 0.887&	0.833 &	0.857 &	0.870 &	0.846 &	0.881 &	0.831 &	0.886 &  - &  -  &	0.867 &	0.860 &	0.854 &	0.861 &	0.829 &	0.848 &	\textbf{0.892}  \cr
			
			&\textbf{$MAE\downarrow$ }          
			& 0.036&	0.050 &	0.040 &	0.038  &	0.039 &	0.035 &	0.049 &	0.037 &	0.035& 	0.038 &	0.043 &	0.040 &	0.041 &	0.042 &	0.054 &	0.043 &\textbf{	0.032} \cr
			
			\hline
			\multirow{5}{*}{\textbf{\rotatebox{270}{RGBD135}}}
			&\textbf{$adpEm\uparrow$}    
			& 0.957& 0.925 &	0.969 &	0.970  &	0.972 &	0.983 &	0.969 &	0.960 &  - & 0.978 &	0.922 &	0.972 &	0.956 &	0.952 &	0.970 &	0.964 &	\textbf{0.979 }\cr
			
			&\textbf{$Sm\uparrow$}       
			& 0.916& 0.861 &	0.930 &	0.937  &	0.951 &	0.946 &	0.927 &	0.918 &	0.931 & 0.939 &	0.886 &	0.933 &	0.910 &	0.919 &	0.925 &	0.922 &\textbf{	0.941} \cr
			
			&\textbf{$adpFm\uparrow$}  
			& 0.898& 0.830 &	0.900 &	0.909  &	-  & 0.936 &	0.907 &	0.895 &	0.916 &	0.939 &	0.865 &	0.907 &	0.891 &	0.875 &	0.910 &	0.909 &	\textbf{0.937} \cr
			
			&\textbf{$WF\uparrow$}  
			& 0.890& 0.774 &	0.894 &	0.896 & 0.915 &	0.932 &	0.884 &	0.884 &	- & -  &	0.845 &	0.902 &	0.863 &	0.877 &	0.900 &	0.887 &	\textbf{0.922} \cr
			
			&\textbf{$MAE\downarrow$ }          
			& 0.021& 0.038 &	0.020 &	0.020  &	0.016 &	0.014 &	0.022 &	0.022 &	0.018 &	0.017 &	0.028 &	0.019 &	0.025 &	0.024 &	0.021 &	0.022 &	\textbf{0.015} \cr
			
			\bottomrule[2pt]
		\end{tabular}}

\end{table*}

\subsection{\textbf{Hybrid Loss Function}}
We impose the BCE loss, IOU loss, and SSIM loss to jointly train our model. The BCE loss is widely used in binary classification and image segmentation. The IOU loss emphasizes the large foreground regions to improve the performance of large objects. The SSIM loss measures the structural similarity between prediction and ground truth. We combine the advantages of the three different loss functions to structure a hybrid loss: 
\begin{equation}
	\mathcal{L}_{H} = \mathcal{L}_{BCE} + \mathcal{L}_{IOU} + \mathcal{L}_{SSIM} ,
\end{equation}
where $\mathcal{L}_{H}$, $\mathcal{L}_{BCE}$, $\mathcal{L}_{IOU}$ and $\mathcal{L}_{SSIM}$ represent the hybrid loss, BCE loss, IOU loss and SSIM loss, respectively. Concretely, our model have three supervision branches, including SFP, TFP and Decoder. Therefore, the total loss is described as:
\begin{equation}
	\begin{split}
	\mathcal{L}_{total} = \mathcal{L}_{H}(S_p, G_{s}) +\mathcal{L}_{H}(T_p, G_{e})+ \mathcal{L}_{H}(S, G_{s}),
	\end{split}
\end{equation}
where $S_p$ and $T_p$ are the output results of SFP and TFP via a $1 \times 1$ Convolution and sigmoid activation function. $S$ is the final saliency prediction of our model. $G_{s}$ and $G_{e}$ indicate the saliency GT and edge GT.

\begin{figure*}
	\centering
	\begin{minipage}{0.30\linewidth}
		\centering
		\includegraphics[width=0.95\linewidth]{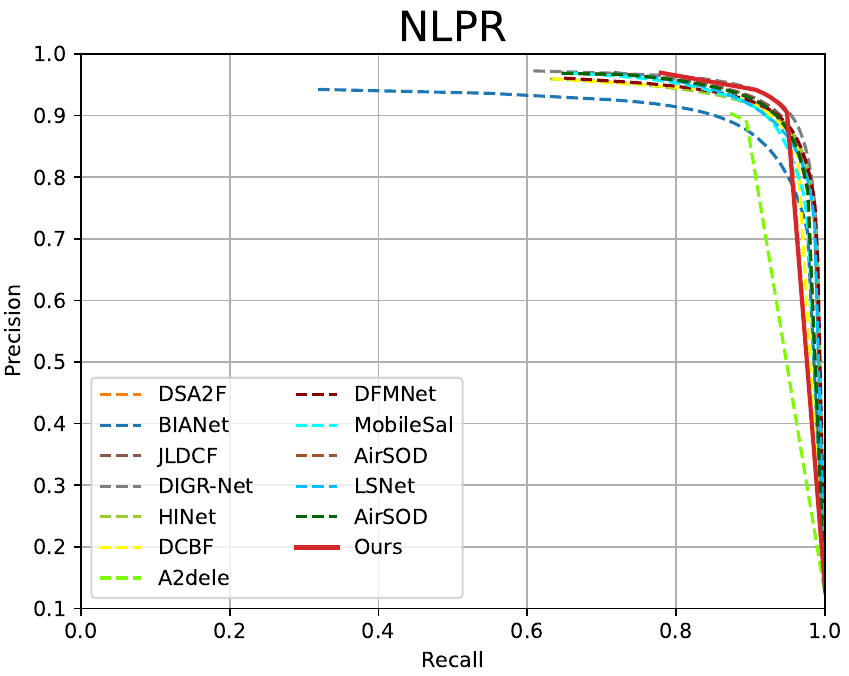}
	\end{minipage}
	\begin{minipage}{0.30\linewidth}
		\centering
		\includegraphics[width=0.95\linewidth]{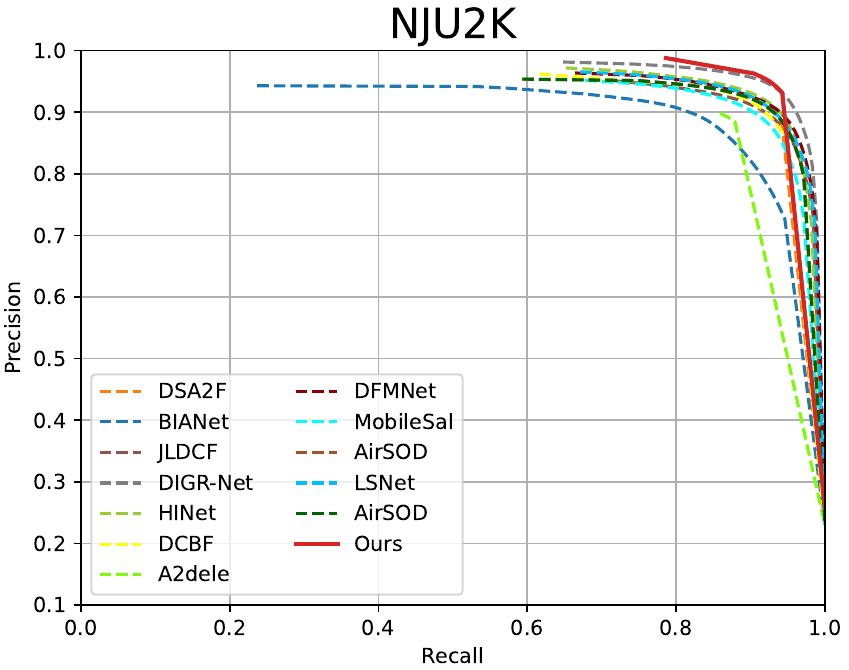}
	\end{minipage}
	\begin{minipage}{0.30\linewidth}
		\centering
		\includegraphics[width=0.95\linewidth]{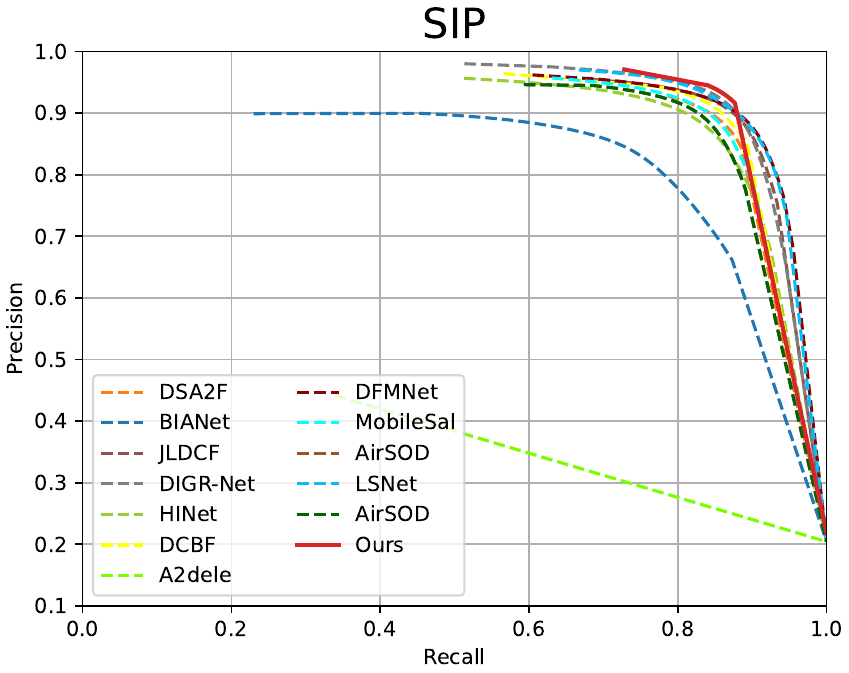}
	\end{minipage}
	
	\begin{minipage}{0.30\linewidth}
		\centering
		\includegraphics[width=0.95\linewidth]{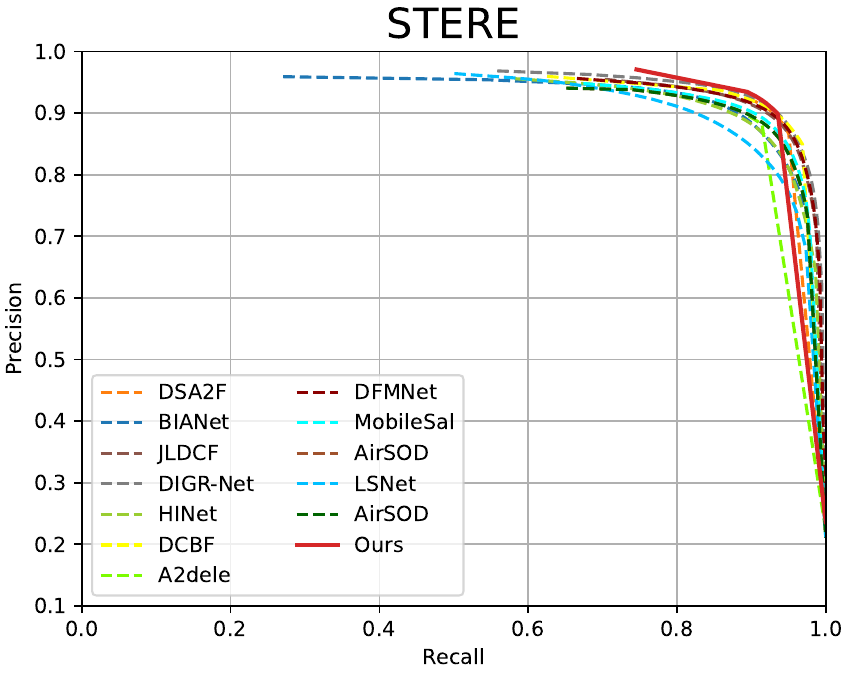}
	\end{minipage}
	\begin{minipage}{0.30\linewidth}
		\centering
		\includegraphics[width=0.95\linewidth]{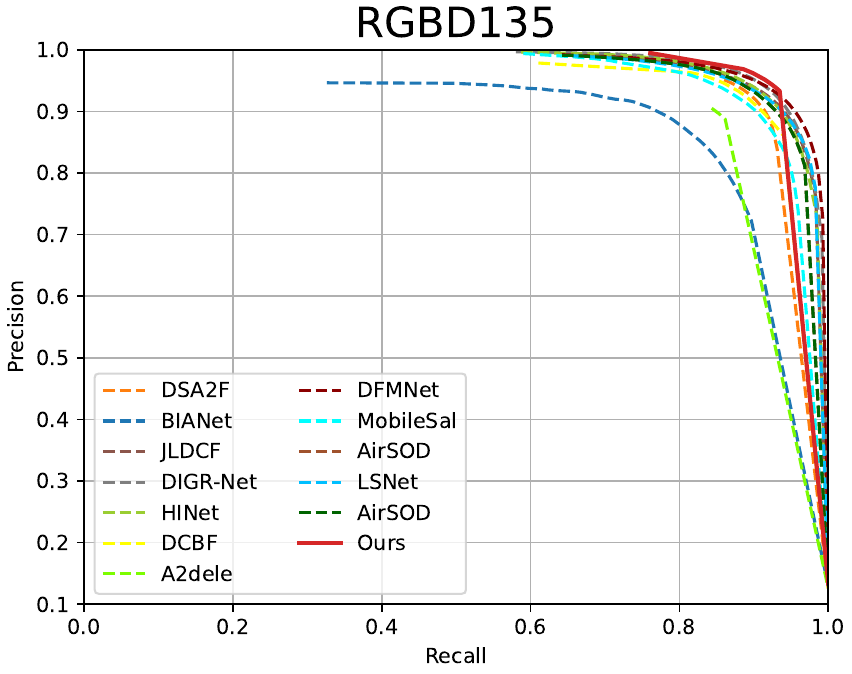}
	\end{minipage}
	
	\caption{Precision-Recall curves of our model and other RGB-D SOD methods on NLPR, NJU2K, SIP, STERE, and RGBD135 datasets.}
	\label{Fig.7}
\end{figure*}

\section{\textbf{Experiments}}

\subsection{\textbf{Datasets}}
We conduct comprehensive experiments and evaluations on 5 available benchmark datasets, like SIP \cite{ref-65},  NJU2K \cite{ref-66}, STERE \cite{ref-68}, RGBD135 \cite{ref-65}, and NLPR \cite{ref-67}. RGBD135 is a pint-sized dataset, which only contains 135 simple indoor images and is adopted for inference. NLPR is a medium-scale dataset, which contains 1000 indoor and outdoor RGB-D images. The NLPR is usually divided into a training set (700 samples) and a testing set (300 samples). SIP is a new and specific dataset with 929 samples and high-quality depth maps, which focus on human-centered scenes to explore the visual bias of human attention mechanisms. NJU2K is collected from the Internet, and 3D movies, which result in terrible depth maps. In this dataset, 1485 pairs are selected as the training set, and 500 pairs are tested for evaluation. STERE is collected from existing stereo datasets, which contains many vivid RGB images and disordered depth maps.


\subsection{\textbf{Evaluation metrics}}
To quantitatively evaluate the RGB-D SOD approaches, we use widely accepted metrics, including Precision-Recall (PR) curve, adapt E-measure ($adpEm$) \cite{ref-69}, S-measure ($Sm$) \cite{ref-70}, adapt F-measure ($adpFm$) \cite{ref-71}, Weighted F-measure ($WF$)) \cite{ref-72}, and Mean Absolute Error ($MAE$) \cite{ref-45}. $Sm$ calculates the similarity of predict and GT saliency maps from a structural perspective. Similar to $Sm$,  $adpEm$ computes similarity from the perspective of global statistics and local matching. $MAE$ measures the error per pixel.

\begin{table}	
	\normalsize
	\setlength\tabcolsep{1pt}
	\centering
	\caption{\label{comparison} Quantitative comparisons of SATNet with Swin-T backbones against the other Transformer-based methods. $\uparrow$/$\downarrow$ indicates that a larger/smaller is better.}
	\label{tab:distortion_type}
	\renewcommand\arraystretch{1.0}
	\resizebox{\linewidth}{!}{
		\begin{tabular}{cc|cccccccc}
			\toprule[1.5pt]
			
			\multicolumn{2}{c|}{\textbf{Method }} &
			\multicolumn{1}{c}{\textbf{PICRNet }} &
			\multicolumn{1}{c}{\textbf{SPNet}} &
			\multicolumn{1}{c}{\textbf{CVAER}} &
			\multicolumn{1}{c}{\textbf{MITF}} &
			\multicolumn{1}{c}{\textbf{CATNet}} &
			\multicolumn{1}{c}{\textbf{EM-Trans}} &
			\multicolumn{1}{c}{\textbf{HFMD}} &
			\multirow{3}{*}{\makecell[c]{\textbf{\textit{SATNet}} \\ \textbf{Ours}}}
			\cr
			
			\multicolumn{2}{c|}{\textbf{Pub. Year }}	
			& 2023 & 2023 &	2023 &	2023 & 2024	&2024 & 2024 &  \cr
			
			\multicolumn{2}{c|}{\textbf{Publication }} 
			& MM & MM & TIP & TCSVT & TMM & TNNLS & TIM &  \cr
			
			\hline    
			\multicolumn{2}{c|}{\textbf{Params(M)}}	 & 112  & 110   & 93.8 & 127.5 & 262.6	  & -  & 431.6 & 59.4	 \cr  
			\multicolumn{2}{c|}{\textbf{ FLOPs(G) }} & 27.1 & 67.8  & 63.9 & 24.1 & 341.8	  & - & 242.2 & 9.3	 \cr  
			
			\hline
			
			\multirow{5}{*}{\textbf{\rotatebox{270}{NLPR}}}
			&\textbf{$adpEm\uparrow$}    
			& 0.968 &	0.970 &	0.961 & 0.963 &	0.971 &	0.968 & 0.968 &	0.967 \cr
			
			&\textbf{$Sm\uparrow$}       
			& 0.935 &	0.939 &	0.929 & 0.933 &	0.940 &	0.939 & 0.938 &	0.930 \cr
			
			&\textbf{$adpFm\uparrow$}  
			& 0.911 &	0.924 &	0.901 & 0.908 &	0.917 &	0.919 & 0.924 &	0.916 \cr
			
			&\textbf{$WF\uparrow$}  
			& 0.907 &	0.917 &	0.895 & - &	0.912 &	0.913 & - &	0.907 \cr
			
			&\textbf{$MAE\downarrow$ }          
			& 0.019 &	0.016 &	0.022 & 0.018 &	0.018 &	0.017 & 0.017 &	0.019 \cr
			\hline
			
			\multirow{5}{*}{\textbf{\rotatebox{270}{NJU2K}}}
			&\textbf{$adpEm\uparrow$}    
			& 0.950 &	0.959 &	0.951 & 0.928 &	0.958 &	0.956 & 0.939 &	0.959 \cr
			
			&\textbf{$Sm\uparrow$}       
			& 0.927 &	0.934 &	0.920 & 0.923 &	0.932 &	0.931 & 0.937 &	0.925 \cr
			
			&\textbf{$adpFm\uparrow$}  
			& 0.919 &	0.930 &	0.913 & 0.917 &	0.929 &	0.926 & 0.937 &	0.927 \cr
			
			&\textbf{$WF\uparrow$}  
			& 0.908 &	0.921 &	0.900 & - &	0.919 &	0.915 & - &	0.916 \cr
			
			&\textbf{$MAE\downarrow$ }          
			& 0.029 &	0.024 &	0.031 & 0.030 &	0.026 &	0.027 & 0.023 &	0.026 \cr
			\hline
			
			\multirow{5}{*}{\textbf{\rotatebox{270}{SIP}}}
			&\textbf{$adpEm\uparrow$}    
			& 0.937 &	0.958 &	0.933 & 0.933 &	0.948 &	0.941 & 0.925 &	0.935 \cr
			
			&\textbf{$Sm\uparrow$}       
			& 0.899 &	0.919 &	0.893 & 0.899 &	0.911 &	0.904 & 0.886 &	0.900 \cr
			
			&\textbf{$adpFm\uparrow$}  
			& 0.900 &	0.925 &	0.892 & 0.903 &	0.912 &	0.912 & 0.897 &	0.913 \cr
			
			&\textbf{$WF\uparrow$}  
			& 0.876 &	0.910 &	0.868 & - &	0.890 &	0.886 & - &	0.887 \cr
			
			&\textbf{$MAE\downarrow$ }          
			& 0.040 &	0.028 &	0.042 & 0.040 &	0.034 &	0.039 & 0.044 &	0.038 \cr
			\hline
			
			\multirow{5}{*}{\textbf{\rotatebox{270}{STERE}}}
			&\textbf{$adpEm\uparrow$}    
			& 0.952 &	0.958 &	0.949 & 0.922 &	0.948 &	0.955 & 0.933 &	0.958 \cr
			
			&\textbf{$Sm\uparrow$}       
			& 0.920 &	0.922 &	0.914 & 0.909 &	0.911 &	0.925 & 0.918 &	0.923 \cr
			
			&\textbf{$adpFm\uparrow$}  
			& 0.905 &	0.912 &	0.897 & 0.888 &	0.912 &	0.913 & 0.905 &	0.919 \cr
			
			&\textbf{$WF\uparrow$}  
			& 0.892 &	0.902 &	0.883 & - &	0.890 &	0.900 & - &	0.907 \cr
			
			&\textbf{$MAE\downarrow$ }          
			& 0.031 &	0.027 &	0.033 & 0.034 &	0.034 &	0.028 & 0.031 &	0.028 \cr
			\hline
			
			\multirow{5}{*}{\textbf{\rotatebox{270}{RGBD135}}}
			&\textbf{$adpEm\uparrow$}    
			& 0.978 &	0.982 &	0.975 & 0.978 &	0.980 & - & - &	0.976 \cr
			
			&\textbf{$Sm\uparrow$}       
			& 0.942 &	0.947 &	0.934 & 0.938 &	0.945 &	- & - &	0.937 \cr
			
			&\textbf{$adpFm\uparrow$}  
			& 0.930 &	0.938 &	0.922 & 0.925 &	0.923 & - & - &	0.941 \cr
			
			&\textbf{$WF\uparrow$}  
			& 0.920 &	0.934 &	0.910 & - &	0.923 & - & - &	0.928 \cr
			
			&\textbf{$MAE\downarrow$ }          
			& 0.016 &	0.014 &	0.018 & 0.014 &	0.016 &	- & - &	0.014 \cr
			\bottomrule[1.5pt]
			
	\end{tabular}}
\end{table}

\subsection{\textbf{Implementation details}}
Following uniform settings, we pick out 650, 1400, and 800 training samples from the NLPR, NJUD, and DUT datasets, respectively. Several data augmentation technologies are used to enhance the training sample's diversity, like randomly flipping, rotating, and resizing ($256\times 256$). All training and testing are conducted by PyTorch framework with an NVIDIA GTX 4090 GPU. We initialize the parameters of the lightweight backbone MobileNet v2 \cite{ref-21} via the pre-trained parameters on ImageNet. Batch size and learn rate are set to 16 and $1e^{-5}$.


\subsection{\textbf{Comparison with SOTA methods}}
To verify the effectiveness of the our model, we conduct extensive comparisons on five public RGB-D datasets with 16 SOTA methods, including DSA2F \cite{ref-73}, BiANet \cite{ref-13}, JLDCF \cite{ref-14}, DIGR \cite{ref-74}, DGFNet \cite{ref-38}, FCFNet \cite{ref-76}, MPDNet \cite{ref-77}, HiDANet \cite{ref-37}, HINet \cite{ref-78}, DCBF \cite{ref-79}, A2dele \cite{ref-50}, DFMNet \cite{ref-80}, MobileSal \cite{ref-52}, MoADNet \cite{ref-15}, LSNet \cite{ref-51}, and AirSOD \cite{ref-16}. For authenticity, we use public saliency maps directly provided by the original authors or use source codes to generate saliency maps under the default parameters.

 \begin{figure*}
	\centering\includegraphics[width=0.95\textwidth,height=9.1cm]{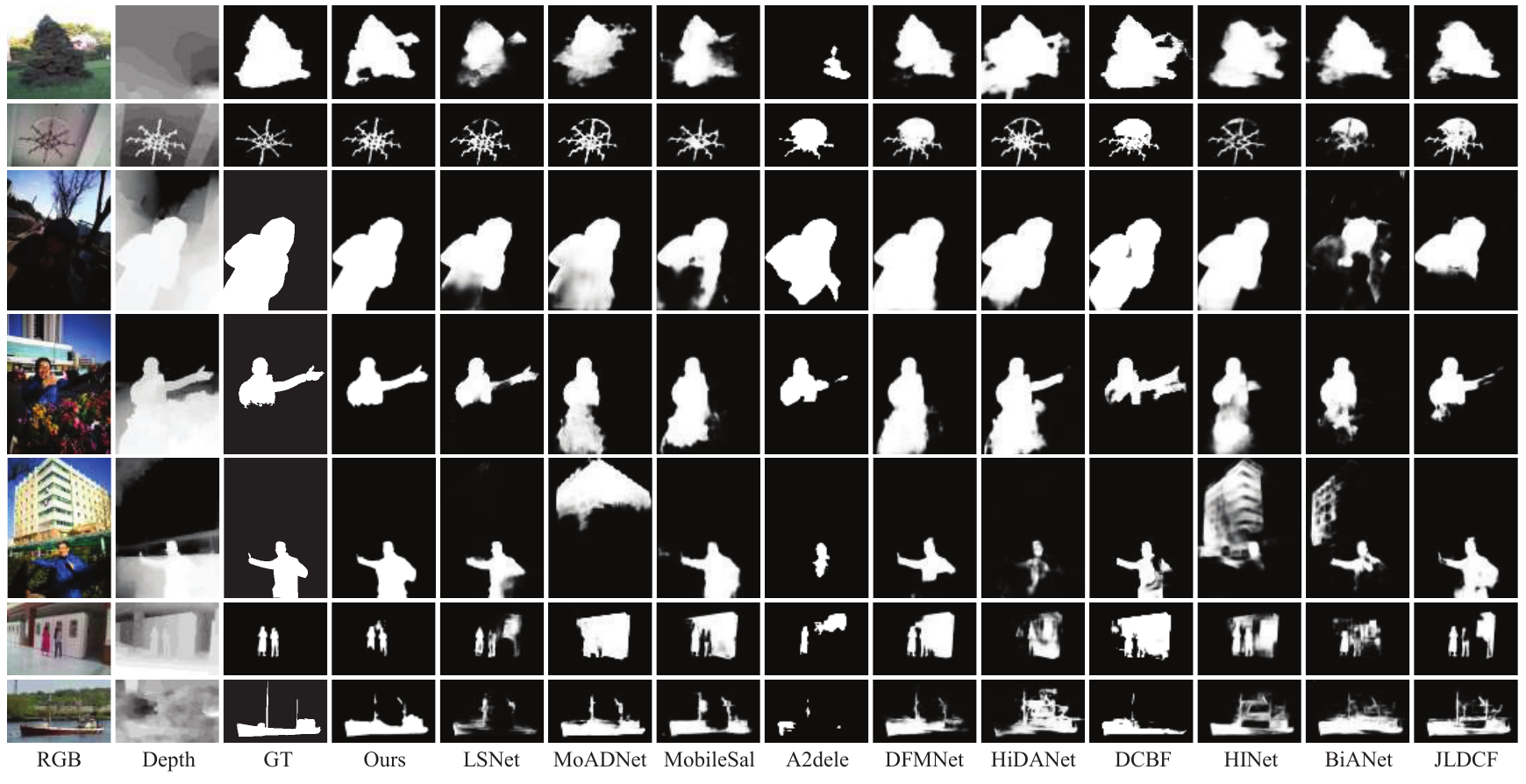}
	\caption{Qualitative comparison of some SOTA RGB-D methods and our model.}
	\label{Fig.8}		
\end{figure*}

 \begin{figure}
	\centering\includegraphics[width=0.48\textwidth,height=6.1cm]{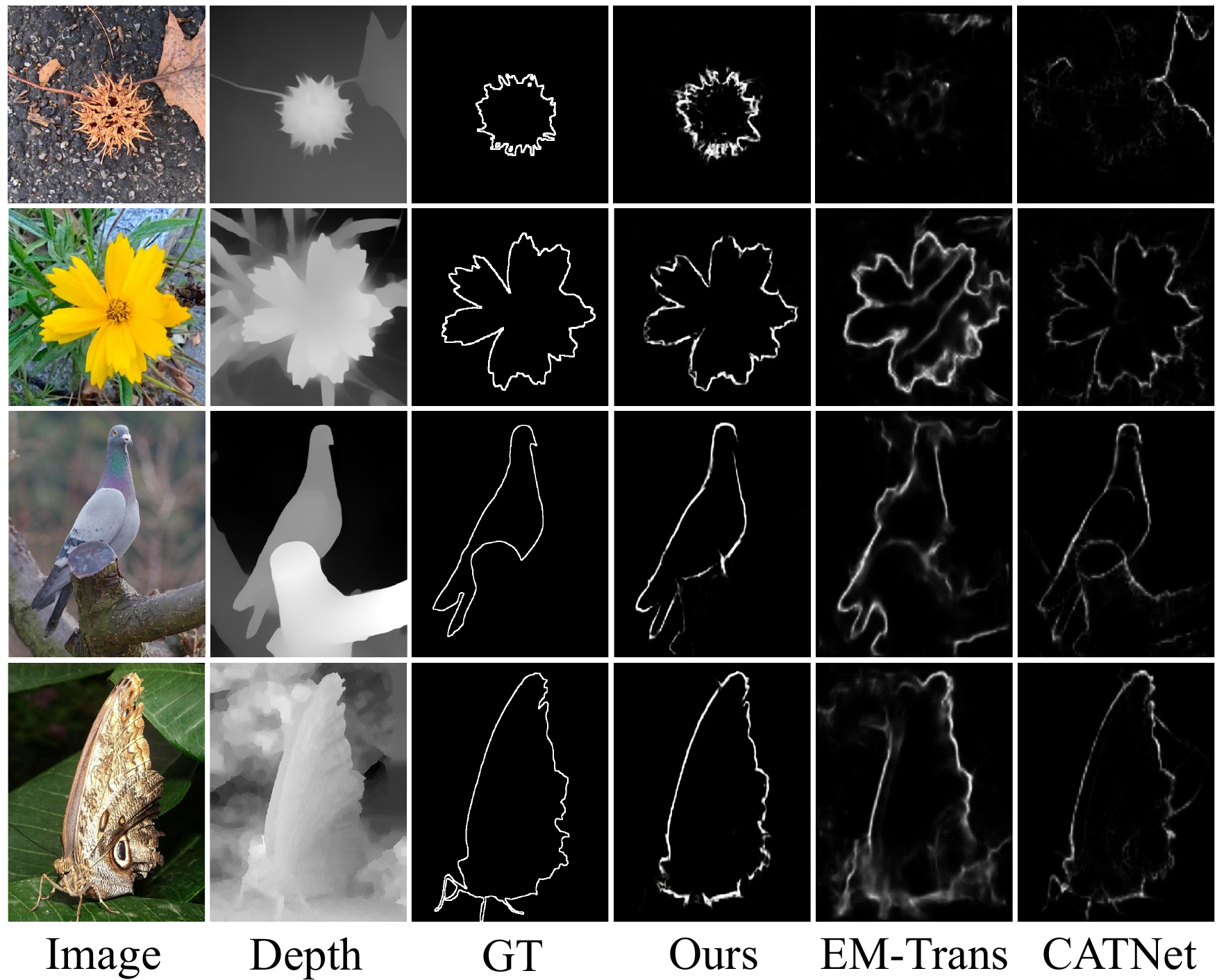}
	\caption{Visual comparison of edge maps, including our SATNet, EM-Trans \cite{ref-87}, and CATNet \cite{ref-86}.}
	\label{Fig.8}		
\end{figure}

 \begin{table}	
	\normalsize
	\setlength\tabcolsep{1pt}
	\setlength{\abovecaptionskip}{-0.01cm}
	\centering
	\caption{\label{comparison} The processing time and parameter of each stage.}
	\label{tab:distortion_type}
	\setlength{\tabcolsep}{10pt} 
	\renewcommand\arraystretch{1}
	\resizebox{\linewidth}{!}{
		\begin{tabular}{c|cccc}
			\toprule[1.2pt]
			\multicolumn{1}{c|}{  Stages } &
			\multicolumn{1}{c}{Encoder } &
			\multicolumn{1}{c}{DAM } &
			\multicolumn{1}{c}{DIRM } &
			\multicolumn{1}{c}{ Decoder} 
			\cr  \hline

			\textit{Times (ms)}             
			& 1.023 &	0.021 &	0.036 &	0.001 
			\cr 
			\textit{Para (M)}           
			& 4.448 & 0.712 &	0.080 &	0.005  
			\cr
			\bottomrule[1.2pt] 
	\end{tabular}} 
\end{table}

\begin{table}[t]	
	\normalsize
	\setlength\tabcolsep{1pt}
	\centering
	\caption{\label{comparison}Effectiveness of the pseudo depth maps on train and test sets. The GD and PD are GT depth maps and pseudo depth maps The best results are highlighted in \textbf{bold}.}
	\label{tab:distortion_type}
	\renewcommand\arraystretch{1}
	\resizebox{\linewidth}{!}{
		\begin{tabular}{c|cc|cc|ccc|ccc}
			\toprule[1.2pt]
			
			\multirow{2}{*}{\textbf{Items}}&
			\multicolumn{2}{c|}{\textbf{  Train Set }} &
			\multicolumn{2}{c|}{\textbf{  Test Set }} &
			\multicolumn{3}{c|}{\textbf{  SIP }} &
			\multicolumn{3}{c}{\textbf{ NLPR}} 
			\cr 
			
			& \textit{GD} & \textit{PD} & \textit{GD} & \textit{PD} &\textbf{$Sm\uparrow$}& \textbf{$adpFm\uparrow$} &\textbf{$MAE\downarrow$ } & \textbf{$Sm\uparrow$}& \textbf{$adpFm\uparrow$} &\textbf{$MAE\downarrow$ } 
			
			\cr \hline

			\textbf{(a)} & \checkmark & & \checkmark &           
			& 0.883 &	0.879 &	0.048 &	0.918 &	0.886 &	0.023 \cr
			
			\textbf{(b)} & \checkmark & &  &   \checkmark     
			& 0.889 &	0.885 &	0.046 &	0.928 &	0.896 &	0.020 \cr 
			
			\textbf{(c)} &  &\checkmark & \checkmark &           
			& 0.887 &	0.881 &	0.046 &	0.921 &	0.893 &	0.021 \cr
			
			\textbf{(d)} &  & \checkmark &  &  \checkmark
			& \textbf{0.894} &	\textbf{0.896} &	\textbf{0.044} &	\textbf{0.932} &	\textbf{0.900} &	\textbf{0.019} \cr
			
			\bottomrule[1.2pt] 
	\end{tabular}} 
\end{table}

 \subsubsection{\textbf{Quantitative comparison}}
 Table \uppercase\expandafter{\romannumeral1} shows the quantitative comparison results on the five datasets with standard metrics. Specifically, the proposed SATNet outperforms other lightweight RGB-D SOD methods on the five datasets with nearly parameters and FLOPs. As reported in Table \uppercase\expandafter{\romannumeral1}, SATNet gets the fastest inference speed compared to lightweight ones, like Ours \textit{vs} AirSOD: 365 \textit{vs} 415. In addition, from the perspective of accuracy comparison, our method still achieve greater advantages. Our SATNet gets 20.8\%, 23.7\%, 10.2\%, 40.7\%, and 28.6\% performance gains than the previous SOTA lightweight method LSNet \cite{ref-51} in terms of \textit{MAE} metric on NLPR, NJU2K, SIP, STERE, and RGBD135. More importantly, our SATNet outperforms CNN-based heavyweight methods. For instance, our SATNet exceeds cutting-edge HiDANet \cite{ref-37} on NLPR, SIP, STERE, and RGBD135, while our parameters and FLOPs are only 4.0\% and 2.1\% of theirs, respectively. These comparison results demonstrate that our SATNet successfully balances performance and efficiency. We also provide the comparisons of the PR curves in Fig. 7. When the recall rate is lower, our precision is higher than other methods, demonstrating that our method has better balance. Note that the disappearance of the upper-left corner of the PR curve for most methods is due to the difficulty of getting (0,1) or (1,0) for (Precision, Recall) at thresholds of 0-255 as perfect overlap between predicted saliency maps and GT is hard to achieve in a dataset.
 
 \begin{table}	
 	\normalsize
 	\setlength\tabcolsep{2pt}
 	\centering
 	\caption{\label{comparison} The performances of MobileSal, AirSOD, and our SATNet with/without pseudo depth maps (PDM). The best results are highlighted in \textbf{bold}.}
 	\label{tab:distortion_type}
 	\renewcommand\arraystretch{1.1}
 	\resizebox{\linewidth}{!}{
 		\begin{tabular}{c|c|ccc|ccc}
 			\toprule[1.2pt]
 			
 			\multirow{2}{*}{\textbf{Items}}&
 			\multirow{2}{*}{\textbf{PDM}}&
 			\multicolumn{3}{c|}{\textbf{  SIP }} &
 			\multicolumn{3}{c}{\textbf{ NLPR}} 
 			\cr 
 			
 			& &\textbf{$Sm\uparrow$}& \textbf{$adpFm\uparrow$} &\textbf{$MAE\downarrow$ } &
 			\textbf{$Sm\uparrow$}& \textbf{$adpFm\uparrow$} &\textbf{$MAE\downarrow$ } 
 			
 			\cr \hline

 			MobileSal  & \ding{55}  & 0.866 & 0.859	& 0.057	& 0.919  & 0.878 & 0.024 \cr
 			
 			MobileSal  & \checkmark & 0.877 & 0.871	& 0.052	& 0.921  & 0.881 & 0.022 \cr
 			
 			AirSOD     & \ding{55}  & 0.859 & 0.855	& 0.060 & 0.925  & 0.884  & 0.023 \cr
 			
 			AirSOD	   & \checkmark & 0.871 & 0.869	& 0.054	& 0.929  & 0.891  & 0.021 \cr
 			
 			SATNet 	& \ding{55} & 0.883 & 0.879	& 0.048 & 0.918  & 0.886  & 0.023 \cr
 			
 			SATNet 	& \checkmark & \textbf{0.894} &	\textbf{0.896} & \textbf{0.044}  & \textbf{0.932} &\textbf{0.900} & \textbf{0.019}	\cr

 			\bottomrule[1.2pt] 
 	\end{tabular}} 
 \end{table}

 \begin{table}[t]	
 	\normalsize
 	\setlength\tabcolsep{1pt}
 	\centering
 	\caption{\label{comparison} Effectiveness of the proposed DAM. The best results are highlighted in \textbf{bold}.}
 	\label{tab:distortion_type}
 	\renewcommand\arraystretch{1.1}
 	\resizebox{\linewidth}{!}{
 		\begin{tabular}{c|ccc|ccc}
 			\toprule[1.2pt]
 			
 			\multirow{2}{*}{\textbf{Variants}}&
 			\multicolumn{3}{c|}{\textbf{  SIP }} &
 			\multicolumn{3}{c}{\textbf{ NLPR}} 
 			\cr 
 			
 			&\textbf{$Sm\uparrow$}& \textbf{$adpFm\uparrow$} &\textbf{$MAE\downarrow$ } &
 			\textbf{$Sm\uparrow$}& \textbf{$adpFm\uparrow$} &\textbf{$MAE\downarrow$ } 
 			
 			\cr \hline

 			\textit{w/o DAM}             
 			& 0.853 &	0.851 &	0.057 &	0.891 &	0.865 &	0.028
 			\cr \hline
 			\textit{w CA}           
 			& 0.877 &	0.874 &	0.052 &	0.919 &	0.889 &	0.025 
 			\cr
 			\textit{w SA}            
 			& 0.876 &	0.880 &	0.052 &	0.921 &	0.887 &	0.024 
 			\cr 
 			\textit{w CBAM}            
 			& 0.875 &	0.871 &	0.053 &	0.920 &	0.895 &	0.023
 			\cr
 			\textit{w SelfA}            
 			& 0.879 &	0.884 &	0.050 &	0.921 &	0.892 &	0.024
 			\cr \hline
 			
 			\textit{Ours-Conv3} & 0.892 & 0.893 & 0.045 & 0.045 & 0.899 & 0.021 \cr 
 			
 			\textit{\textbf{Ours}}  				
 			& \textbf{0.894} & \textbf{0.896} & \textbf{0.044} & \textbf{0.932} & \textbf{0.900} &	\textbf{0.019}
 			\cr
 			
 			\bottomrule[1.2pt] 
 	\end{tabular}} 
 \end{table}

 \subsubsection{\textbf{Qualitative comparison}}
  Fig. 8 presents the visual comparison of the proposed SATNet and other methods, including heavyweight methods and lightweight methods. The saliency maps of the proposed SATNet have more accurate localization and fine-grained edges in several challenging scenes, including low contrast, complex backgrounds, small objects, and poor depth maps. SATNet effectively eliminates background interference and accurately segments foreground objects when processing low-quality depth maps, as shown in $7^{th}$ rows of Fig. 8. In a word, compared with these methods, the proposed SATNet completely segments the foreground objects with sharp edges. Besides, we show the visual comparison of edge maps with EM-Trans \cite{ref-87} and CATNet \cite{ref-86} in Fig. 9. Thanks to the dual modeling of texture and saliency information, our SATNet generates sharper edge maps than EM-Trans and CATNet. For example, our SATNet's edge maps are more complete than the intermittent edge maps generated by EM-Trans and CATNet in the bottom row of Fig. 9.

   \begin{figure}
	\centering\includegraphics[width=0.49\textwidth,height=7cm]{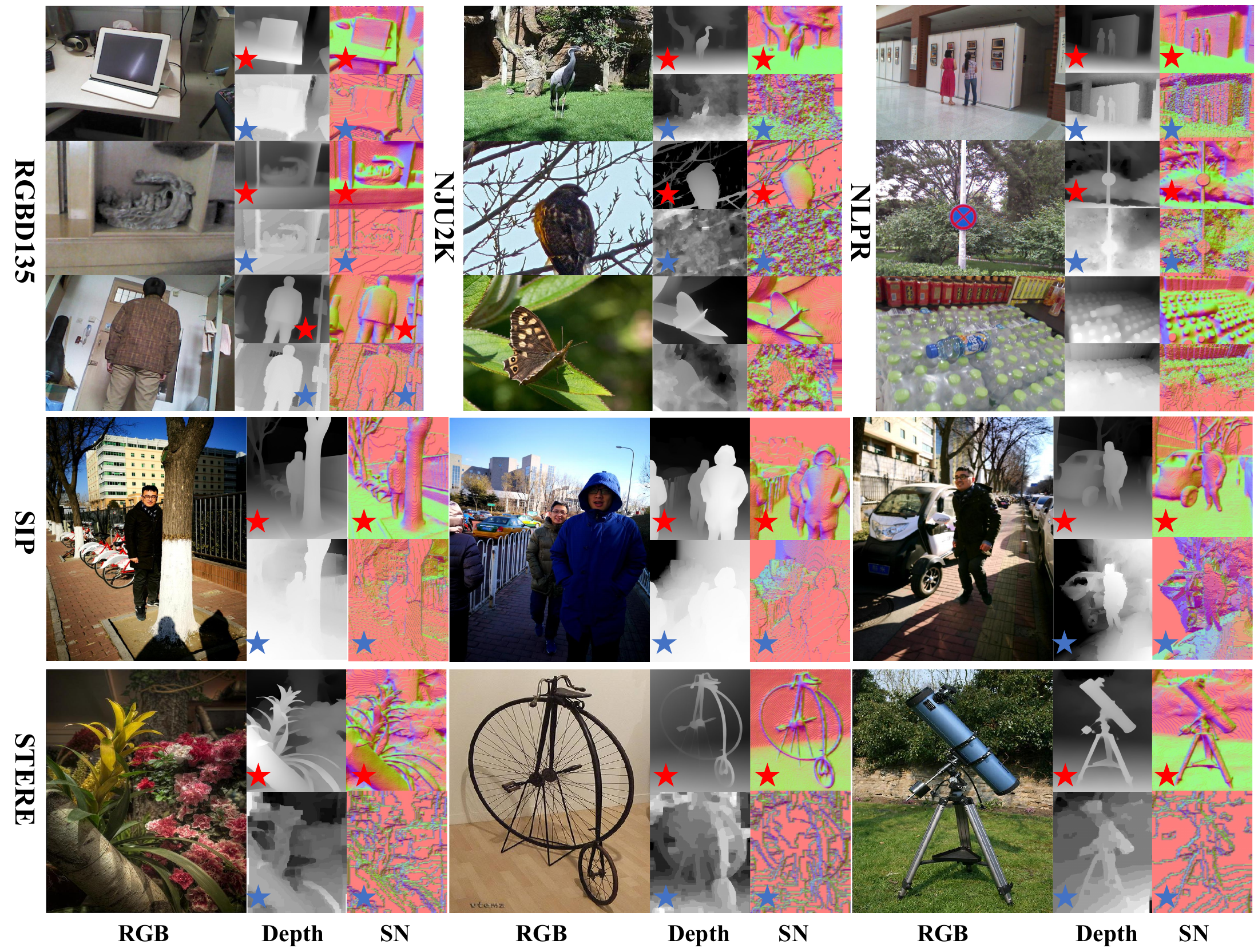}
	\caption{The visual comparison of surface normals (SN). Surface normals are derived from pseudo depth maps and GT depth maps on RGBD135, NJU2K, NLPR, SIP, and STERE datasets. Pseudo depth maps and their corresponding surface normals are denoted by \textcolor{red}{\textbf{red stars}}, while GT depth maps and their associated surface normals are represented by \textcolor[RGB]{68,114,196}{\textbf{blue stars}}.}
	\label{Fig.9}		
\end{figure}

 \subsubsection{\textbf{Compared with Transformer-based methods}}

In recent years, vision transformers have attracted a lot of attention, some works employ vision transformers, like T2T \cite{ref-83} and Swin Transformer \cite{ref-54}, as backbones to extract multi-modality features. To prove the effectiveness of SATNet in the transformer framework, we use the lightweight Swin-tiny Transformer as backbone to rebuild SATNet. Then, we conduct comparison experiments on five benchmarks with five SOTA transformer-based methods, including PICRNet \cite{ref-84}, SPNet \cite{ref-85}, CVAER \cite{ref-88}, CATNet \cite{ref-86}, \added{HFMD \cite{ref-95}}, \added{MITF \cite{ref-96}} and EM-Trans \cite{ref-87}. The comparison results are reported in Table \uppercase\expandafter{\romannumeral2}. For example, SATNet outperforms HFMT's accuracy with only 13.8\% of HFMT's parameters and 3.8\% FLOPs on large-scale SIP and STERE datasets. Besides, SATNet can also achieve very close performance to HFMT on NLPR and NJU2K datasets. Compared with the MITF, our SATNet still better segmentation results with few parameters and FLOPs. Compared to PICRNet, the lightest of these methods, SATNet outperforms PICRNet with only half the parameters and one-third of the FLOPs, which proves the effectiveness and efficiency of SATNet.

 \subsubsection{\textbf{Efficiency comparison}}
 In addition to the performance comparison, we also make an efficiency comparison, as shown in Table \uppercase\expandafter{\romannumeral1}. We observe that the FPS metrics of the high-complexity RGB-D SOD methods are low because of the high computations and parameters. Our model outperforms these SOTA methods by a large margin from an efficiency view. For the lightweight methods, the performances of these methods are suboptimal compared with SOTA methods. By contrast, our model outperforms other lightweight methods and the heavyweight method (HiDANet) with the fastest inference speed (415 FPS) and 5.2 parameters. Furthermore, we list the processing times and parameters of each stage for SATNet in Table \uppercase\expandafter{\romannumeral3}. The main contributions (DAM and DIRM) accounted for only 14.5\% of the total number of parameters and 5.4\% of the total run time, which demonstrates the proposed DAM and DIRM are efficient modules.

 \subsection{\textbf{Ablation Study}}

 To illustrate the effectiveness of each component of our model, we provide the following ablation studies in terms of $Sm$, $afpFm$, and $MAE$ on SIP and NLPR datasets.

  \subsubsection{\textbf{The effectiveness of pseudo depth maps}}
  We use the pseudo generated by the Depth Anything Model to alleviate the quality of depth maps in existing RGB-D SOD datasets. We leverage GT depth maps and pseudo depth maps to train and test our SATNet, respectively. The comparison results are shown in Table \uppercase\expandafter{\romannumeral4}, which contains four different cases: (a) Depth Anything Model is not used in either the training or testing; (b) Depth Anything Model is only used in testing; (c) Depth Anything Model is only used in training; (d) Depth Anything Model is used in both the training and testing. The results of pseudo depth maps outperform the one of GT depth maps. Specifically, SATNet-(b) and SATNet-(c) obtain better performance than the SATNet-(a), and SATNet-(d) get the best results. Besides, we also show the visual comparison in Fig. 1 (a) and Fig. 10. Obviously, the advantage of pseudo depth map is very significant. To intuitively show the depth map, we utilize it to generate a surface norm map that measures the smoothness of the surface of an object in Fig. 10. We see that the low-quality depth map contains serious Gaussian-like noise and checkerboard pattern. In contrast, the key appearance information such as the contour of the object is better embodied in pseudo depth map. It is seen that the pseudo depth maps are closer to the ground-truth and retain the human contours very well in the middle of the $4^{th}$ line.
  
  To further prove the effectiveness of the pseudo depth map, we re-train two lightweight RGB-D SOD models (AirSOD and MobileSal) on RGB and pseudo depth images, and the results are reported in Table \uppercase\expandafter{\romannumeral5}. AirSOD and MobileSal obtain 8.8\% and 10\% performance gains in terms of MAE on the SIP dataset. The significant improvements indicate that the pseudo depth map is an effective strategy for improving the SOD models. In addition, our SATNet outperforms the MobileSal and AirSOD in the same training and testing using pseudo-depth maps, demonstrating the effectiveness of our SATNet.

 \subsubsection{\textbf{The effectiveness of DAM}}
 To validate the efficiency of the DAM, we conduct a specific analysis of its design. First, to prove the role of DAM for the total framework, we use simple addition operations instead of the DAM to fuse multi-modality features, denoted as \textit{w/o DAM}. As expressed in the introduction, we consider that the current attention mechanism effectively works in heavyweight models, but can't be effectively migrated to lightweight models. To testify this perspective, we use widely used channel attention (CA), spatial attention (SA), hybrid attention (CBAM), and self-attention (SelfA) instead of DAM, denoted as \textit{w CA}, \textit{w SA}, \textit{w CBAM}, and \textit{w SelfA}, respectively. The results of these variants are reported in Table \uppercase\expandafter{\romannumeral6}. DAM contributes 29.5\% and 33.3\% gains in terms of MAE on SIP and NLPR datasets according to the comparison results of \textit{w/o DAM} and Ours. Furthermore, compared with \textit{w CA}, \textit{w SA}, \textit{w CBAM}, and \textit{w SelfA}, our model obtains the best results, which proves the effectiveness of DAM. Besides, we replace the $7 \times 7$ convolution layer with a $3 \times 3$ one in Table \uppercase\expandafter{\romannumeral6}, and the modified model is named \textit{Ours - Conv3}. The experimental results show that the 7×7 convolution operation can effectively improve the performance of SATNet and is significantly superior to the 3×3 convolution operation. The parameter and FLOPs of our SATNet are only slightly greater than those of \textit{Ours-Conv3}, for example, the parameter is 5245.4K vs 5245.2K, and the FLOPs are 1545M vs 1541M. However, the performance gain is significant, which indicates that the $7 \times 7$ convolution layer is valuable.

  \begin{table}	
  	\normalsize
  	\setlength\tabcolsep{1pt}
  	\centering
  	\caption{\label{comparison} Effectiveness of the proposed DIRM. The best results are highlighted in \textbf{bold}.}
  	\label{tab:distortion_type}
  	\renewcommand\arraystretch{1.1}
  	\resizebox{\linewidth}{!}{
  		\begin{tabular}{l|ccc|ccc}
  			\toprule[1.2pt]
  		
  			\multirow{2}{*}{\textbf{Variants}}&
  			\multicolumn{3}{c|}{\textbf{ SIP }} &
  			\multicolumn{3}{c}{\textbf{ NLPR}} 
  			\cr 
  			
  			& \textbf{$Sm\uparrow$}& \textbf{$adpFm\uparrow$} &\textbf{$MAE\downarrow$ } &
  			\textbf{$Sm\uparrow$}& \textbf{$adpFm\uparrow$} &\textbf{$MAE\downarrow$ } 
 
  			\cr \hline  
  			\textit{SATNet (Ours)} & \textbf{0.894} & \textbf{0.896} & \textbf{0.044} & \textbf{0.932} & \textbf{0.900} &	\textbf{0.019} \cr \hline
  			
  			\textit{Ours-LTR}               &	0.874 &	0.868 &	0.051 &	0.922 &	0.892 &	0.023 \cr
  			\textit{Ours-LTR\&TFP}          &	0.879 &	0.881 &	0.048 &	0.920 &	0.887 &	0.025 \cr 
  			\textit{Ours-GSR}	            &	0.874 &	0.867 &	0.052 &	0.920 &	0.889 &	0.026 \cr
  			\textit{Ours-GSR\&SFP}	        &	0.877 &	0.874 &	0.050 &	0.921 &	0.895 &	0.022 \cr 
  			\textit{Ours-DIRM}	&   0.861 &	0.858 &	0.058 &	0.893 &	0.858 &	0.032 \cr
  			\bottomrule[1.2pt] 
  	\end{tabular}} 
  \end{table}

 \begin{table}	
 	\normalsize
 	\setlength\tabcolsep{1pt}
 	\centering
 	\caption{\label{comparison} Effectiveness of DFAM. The best results are highlighted in \textbf{bold}.}
 	\label{tab:distortion_type}
 	\renewcommand\arraystretch{1.1}
 	\resizebox{\linewidth}{!}{
 		\begin{tabular}{c|ccc|ccc}
 			\toprule[1.2pt]
 			
 			\multirow{2}{*}{\textbf{Variants}}&
 			\multicolumn{3}{c|}{\textbf{  SIP }} &
 			\multicolumn{3}{c}{\textbf{ NLPR}} 
 			\cr 
 			
 			&\textbf{$Sm\uparrow$}& \textbf{$adpFm\uparrow$} &\textbf{$MAE\downarrow$ } &
 			\textbf{$Sm\uparrow$}& \textbf{$adpFm\uparrow$} &\textbf{$MAE\downarrow$ } 
 			
 			\cr \hline

 			\textit{w/o AsymmConv}             
 			& 0.878 & 0.877	& 0.050 & 0.921  & 0.895  & 0.024 
 			\cr
 			\textit{w/o DilatedConv}           
 			& 0.880 &	0.881 & 0.046  & 0.923 & 0.897 & 0.023
 			\cr \hline
 			
 			\textit{w ASPP}             
 			& 0.874 & 0.878	& 0.053 & 0.919  & 0.891  & 0.025 
 			\cr
 			\textit{w DenseASPP}           
 			& 0.881 &	0.882 & 0.048  & 0.923 & 0.892 & 0.022
 			\cr \hline
 			
 			\textit{Ours}  				
 			& \textbf{0.894} & \textbf{0.896} & \textbf{0.044} & \textbf{0.932} & \textbf{0.900} &	\textbf{0.019}
 			\cr
 			
 			\bottomrule[1.2pt] 
 	\end{tabular}} 
 \end{table}

 \begin{table}	
 	\normalsize
 	\setlength\tabcolsep{2pt}
 	\centering
 	\caption{\label{comparison}Ablation of hyper-parameter in DFAM, including convolution kernel (CK) and dilated rate (DR). The results of our SATNet are highlighted in \textbf{bold}.}
 	\label{tab:distortion_type}
 	\renewcommand\arraystretch{1.1}
 	\resizebox{\linewidth}{!}{
 		\begin{tabular}{c|c|ccc|ccc}
 			\toprule[1.2pt]
 			
 			\multirow{2}{*}{\textbf{Items}}&
 			\multirow{2}{*}{\textbf{CK\&DR}}&
 			\multicolumn{3}{c|}{\textbf{ SIP}} &
 			\multicolumn{3}{c}{\textbf{NLPR}} 
 			\cr 
 			
 			& &\textbf{$Sm\uparrow$}& \textbf{$adpFm\uparrow$} &\textbf{$MAE\downarrow$ } &
 			\textbf{$Sm\uparrow$}& \textbf{$adpFm\uparrow$} &\textbf{$MAE\downarrow$ } 
 			
 			\cr \hline

 			(a)  	& (1, 3, 5)	        
 			& 0.889 & 0.891	& 0.046	& 0.928  & 0.896 & 0.020 \cr
 			\textbf{(b)}	& \textbf{(3, 5, 7)} 	    
 			& \textbf{0.894} & \textbf{0.896} & \textbf{0.044} & \textbf{0.932} & \textbf{0.900} &	\textbf{0.019} \cr
 			
 			(c) 	& (3, 7, 11) 	         
 			& 0.891 & 0.890	& 0.045 & 0.926 & 0.893 & 0.021 \cr
 			
 			(d) 	& (3, 9, 15) 	         
 			& 0.881 & 0.883 & 0.049 & 0.925 & 0.891 & 0.022	\cr

 			\bottomrule[1.2pt] 
 	\end{tabular}} 
 \end{table} 
 
  \subsubsection{\textbf{The effectiveness of DIRM}}
  We design a Dual Information Representation Module (DIRM) to extract texture and saliency information via TFP and SFP. To testify effectiveness of the DIRM, we formulate several variants by removing certain components. Specifically, we delete the LTR module and GSR module, denoted as \textit{Ours-LTR} and \textit{Ours-GRS}, respectively. To explain TFP, we omit TFP and LTR module, denote \textit{Ours-LTR\&TFP}. Similar to \textit{Ours-LTR\&TFP}, we delete SFP and GRS, denoted as \textit{Ours-GRS\&SFP}. Furthermore, we delete DIRM, denoted as \textit{Ours-DIRM}. The comparison results of these variants are presented in Table \uppercase\expandafter{\romannumeral7}. Compared with \textit{Ours-DIRM}, our model obtains the percentage gain of 31.8\% and 52.4\% in terms of MAE score on SIP and NLPR datasets, respectively, which illustrates the role of coupling inverted pyramid. Compared \textit{Ours-LTR}, our model still get significant performance superiority, \textit{e.g.}, \textit{Ours} vs \textit{Ours-LTR}, MAE: 0.044$\rightarrow$0.051 on SIP and 0.019$\rightarrow$0.023 on NLPR dataset, which prove the effectiveness of TFP and LTR module. Similar to the TFP and LTR, the function of SFP and GSR modules also are proved from Table \uppercase\expandafter{\romannumeral7}.

  \subsubsection{\textbf{The effectiveness of DFAM}}
  We utilize asymmetric convolution and dilated convolution to capture large-scale receptive filed information in DFAM. Therefore, to identify the effectiveness of the DFAM, we remove the asymmetric convolution and dilated convolution, denoted as \textit{w/o AsymmConv} and \textit{w/o DilatedConv}, respectively.  Besides, to prove the superiority of DFAM, we replace it with ASPP \cite{ref-90} and DenseASPP \cite{ref-89}, denoted as \textit{w ASPP} and \textit{w DenseASPP}. The comparison results are reported in Table \uppercase\expandafter{\romannumeral8}. We see that our model outperforms the \textit{w/o AsymmConv} and \textit{w/o DilatedConv}. Besides, our SATNet gets better performance compared with \textit{w ASPP} and \textit{w DenseASPP}, like \textit{Sm}: 0.894 \textit{vs} 0.878 and 0.881, MAE: 0.044 \textit{vs} 0.053 and 0.048 on SIP datasets. Obviously, the combination of asymmetric and dilated convolutions is effective in a lightweight framework. To analyze the role of different convolution kernels and dilated rates, we set different kernel sizes and rates in Table IX. The best performance is same as our SATNet in manuscript. As the CK and DR increases (from (3,5,7) to (3,9,15)), there is a slight decrease in performance for both datasets. The configuration (3,7,11) maintains relatively high performance but does not surpass the (3,5,7) configuration. The results demonstrate that CK and DR parameters is crucial for optimizing the performance of the DFAM within SATNet. The reason why a larger convolution kernel does not necessarily lead to better results is that large asymmetric convolutions can disrupt the local spatial continuity of features, thereby degrading the model's performance.

    \subsubsection{\textbf{The exploration of Efficiency Factor}}
  After DAM, we set an Efficiency Factor ($\psi_{ef}$) to unify the channels of the texture feature pyramid and semantic feature pyramid. The $\psi_{ef}$ reduces the feature spaces of the subsequent model. The $\psi_{ef}$ is set to 32 in our model. To explore the optimal setting, we formulate different variants with different $\psi_{ef}$ (16, 64, and 128), denoted as \textit{EF16}, \textit{EF64}, and \textit{EF128}, where results of these variants are presented in Table \uppercase\expandafter{\romannumeral10}. Although \textit{EF16} achieves fewer FLOPs (1.1 G), their performance degradation is too poor in Table \uppercase\expandafter{\romannumeral10}. Compared with EF16, our model gets a percentage gain of 16\% and 19\% in terms of MAE score on SIP and NLPR with a slight increase of parameters. Moreover, EF64 and EF128 possess more rich channel features and realize a tiny performance improvement compared with our model. However, EF64, and EF128 generate more large FLOPs and Parameters, \textit{e.g.}, Ours$\rightarrow$Ef64 (EF128), FLOPs: 1.5$\rightarrow$3.2 (8.9). We think this operation is unworthy and inadvisable. Therefore, the efficiency factor is set to 32 is a suitable choice.

 \begin{table}	
 	\normalsize
 	\setlength\tabcolsep{1pt}
 	\centering
 	\caption{\label{comparison} Exploration of the Efficiency Factor ($\psi_{ef}$). The results of our model are highlighted in \textbf{bold}.}
 	\label{tab:distortion_type}
 	\renewcommand\arraystretch{0.9}
 	\resizebox{\linewidth}{!}{
 		\begin{tabular}{c|c|ccc|ccc}
 			\toprule[1.2pt]
 			
 			\multirow{2}{*}{\textbf{Variants}}&
 			\multirow{2}{*}{\textbf{FLOPs (G)}}&
 			\multicolumn{3}{c|}{\textbf{  SIP }} &
 			\multicolumn{3}{c}{\textbf{ NLPR}} 
 			\cr 
 			
 			& &\textbf{$Sm\uparrow$}& \textbf{$adpFm\uparrow$} &\textbf{$MAE\downarrow$ } &
 			\textbf{$Sm\uparrow$}& \textbf{$adpFm\uparrow$} &\textbf{$MAE\downarrow$ } 
 			
 			\cr \hline  
 			
 			\textit{EF16}  &  1.1 & 	0.867 &	0.876 &	0.051 &	0.901 &	0.884 &	0.025 \cr
 			\textit{\textbf{Ours}}  & 	\textbf{1.5}	& \textbf{0.894} & \textbf{0.896} & \textbf{0.044} & \textbf{0.932} & \textbf{0.900} &	\textbf{0.019} \cr
 			\textit{EF64}  &  3.2	& 	0.887 &	0.897 &	0.043 &	0.928 &	0.910 &	0.021 \cr
 			\textit{EF128} & 8.9	& 	0.886 &	0.895 &	0.044 &	0.927 &	0.899 &	0.020 \cr
 			
 			\bottomrule[1.2pt] 
 	\end{tabular}} 
 \end{table}

  \begin{figure}[t]
 	\centering\includegraphics[width=0.48\textwidth,height=5cm]{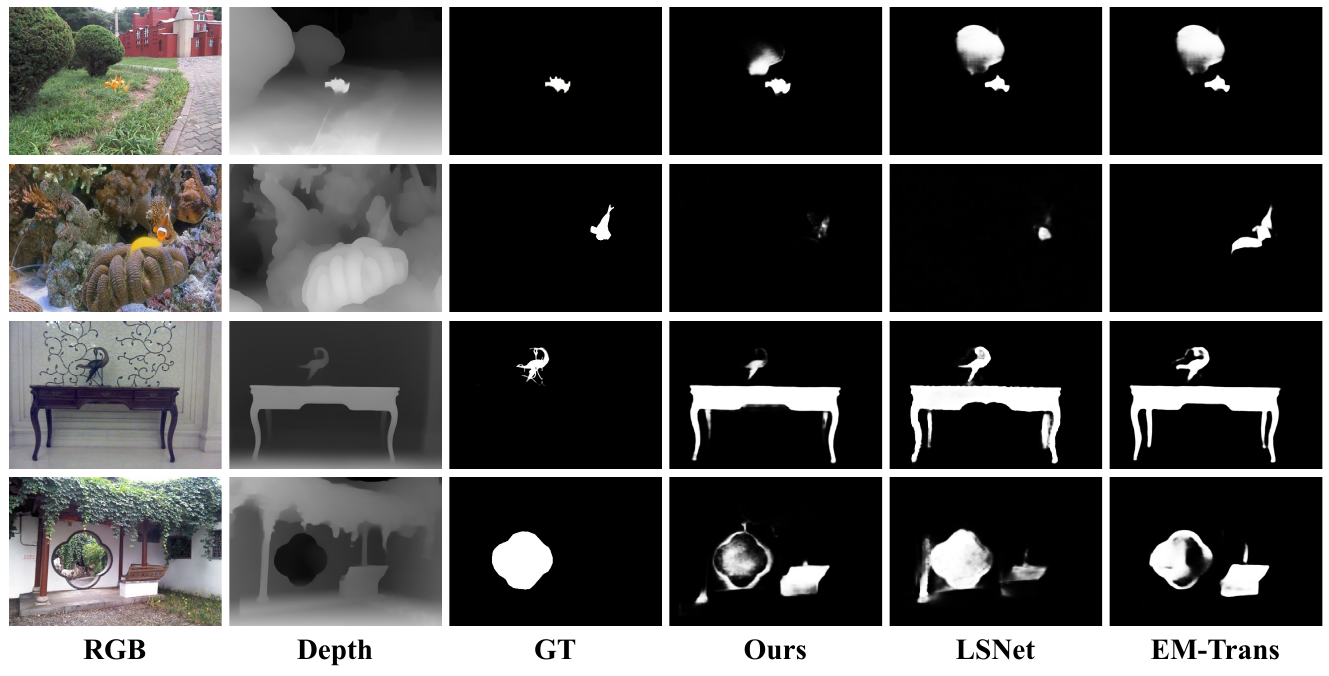}
 	\caption{Some failure cases of our SATNet, LSNet \cite{ref-51} and EM-Trans \cite{ref-87}.}
 	\label{Fig.10}		
 \end{figure}

 \subsection{\textbf{Failure Cases}}
 We have demonstrated the advantages and effectiveness of the proposed SATNet through qualitative and quantitative experiments. However, our SATNet may fail in some extremely challenging scenarios. Fig. 11 shows some failure cases of our SATNet and two recently proposed SOTA models. In the top-two rows, the salient object is too small and background is confused, which misleads detection model and results in terrible results for our SATNet. In the bottom-two rows, the depth distance of the background is greater than or equal to that of the salient object, which may result in the corresponding depth map failing to provide distinguishable depth information between the salient object and the background. Therefore, it is challenging to suppress the background distractors in such a scene. It is noteworthy that even the latest SOTA methods (LSNet \cite{ref-51} and EM-Trans \cite{ref-87}) struggle to achieve accurate segmentation results for the aforementioned challenging scenes.

\subsection{\textbf{Expanded Applications}}
To further demonstrate the effectiveness of our SATNet on other foreground segmentation tasks, we extend our SATNet on RGB-T salient object and polyp segmentation tasks.
 
 \begin{table}	
 	\normalsize
 	\setlength\tabcolsep{1pt}
 	\centering
 	\caption{\label{comparison}The quantitative comparison of SATNet with EfficientNet and Swin-T on polyp segmentation task. SATNet-E/S are the results of our SATNet with EfficientNet and Swin-T backbones.}
 	\label{tab:distortion_type}
 	\renewcommand\arraystretch{0.9}
 	\resizebox{\linewidth}{!}{
 		\begin{tabular}{c|c|cc|cc|cc}
 			\toprule[1.2pt]
 			
 			\multirow{2}{*}{\textbf{Methods}}&
 			\multirow{1}{*}{\textbf{FLOPs}}&
 			\multicolumn{2}{c|}{\textbf{CVC-300}} &
 			\multicolumn{2}{c|}{\textbf{ColonDB}} &
 			\multicolumn{2}{c}{\textbf{ETIS}} 
 			\cr 
 			
 			& (G) & \textbf{$adpFm\uparrow$} & \textbf{$MAE\downarrow$ } & \textbf{$adpFm\uparrow$} & \textbf{$MAE\downarrow$ }  & \textbf{$adpFm\uparrow$} &\textbf{$MAE\downarrow$ }  	\cr \hline  
 			
 			PraNet  	& 32.6 	& 0.824 & 0.010 &	0.718 &	0.043 &	0.602 &	0.031  \cr
 			CTNet 		& 15.2  & 0.885 & 0.006 &	0.781 &	0.027 &	0.723 &	0.014 \cr
 			SEPNet 		& 12.5	& 0.862 &	0.006 &	0.768 &	0.027 &	0.724 &	0.016  \cr
 			BUNet 	    & 10.7  & 0.081 &	0.007 &	0.758 &	0.029 &	0.717 &	0.017  \cr
 			CGLCON 	    & 169.1 & 0.840 &	0.008 &	0.766 &	0.030 &	0.707 &	0.016  \cr
 			SATNet-E 	& 1.5 	& 0.869 &	0.006 &	0.725 &	0.036 &	0.679 &	0.015  \cr
 			SATNet-S 	& 9.3 	& 0.889 & 	0.005 &	0.765 &	0.030 &	0.725 &	0.014  \cr 
 			\bottomrule[1.2pt] 
 	\end{tabular}} 
 \end{table}

 \begin{table}	
 	\normalsize
 	\setlength\tabcolsep{1pt}
 	\centering
 	\caption{\label{comparison}The quantitative comparison of SATNet with EfficientNet and Swin-T on RGB-T SOD task. SATNet-E/S are the results of our SATNet with EfficientNet and Swin-T backbones.}
 	\label{tab:distortion_type}
 	\renewcommand\arraystretch{0.9}
 	\resizebox{\linewidth}{!}{
 		\begin{tabular}{c|c|cc|cc|cc}
 			\toprule[1.2pt]
 			\multirow{2}{*}{\textbf{Methods}}&
 			\multirow{1}{*}{\textbf{FLOPs}}&
 			\multicolumn{2}{c|}{\textbf{VT821}} &
 			\multicolumn{2}{c|}{\textbf{VT1000}} &
 			\multicolumn{2}{c}{\textbf{VT5000}} 
 			\cr 
 			
 			& (G) & \textbf{$adpFm\uparrow$} & \textbf{$MAE\downarrow$ } & \textbf{$adpFm\uparrow$} & \textbf{$MAE\downarrow$ }  & \textbf{$adpFm\uparrow$} &\textbf{$MAE\downarrow$ }  	\cr \hline

 			CGFNet		&	231.1	&	0.821 &	0.038 &	0.901 &	0.023 &	0.835 &	0.035 \cr
 			ECFFNet		&	-	 	&   0.809 &	0.035 &	0.876 &	0.022 &	0.807 &	0.038 \cr
 			SwinNet		&   124.3	&   0.836 &	0.030 &	0.896 &	0.018 &	0.861 &	0.026 \cr
 			TriTransNet	&	292.3	&	0.843 &	0.026 &	0.899 &	0.017 &	0.851 &	0.031 \cr 
 			XMSNet		&   -	    &   0.845 &	0.028 &	0.902 &	0.018 &	0.865 &	0.028 \cr
 			SATNet-E	&	1.5		&	0.811 &	0.034 &	0.891 &	0.023 &	0.830 &	0.035 \cr
 			SATNet-S	&   9.3		&	0.847 &	0.027 &	0.905 &	0.018 &	0.856 &	0.030 \cr	
 			\bottomrule[1.2pt] 
 	\end{tabular}} 
 \end{table} 
 
\subsubsection{Polyp Segmentation} 
Polyp segmentation is important for the diagnosis and treatment of colon cancer. Similar to the SOD, polyp segmentation aims to segment the foreground objects, which are the polyp regions. We train our SATNet on polyp segmentation datasets, including CVC-300 \cite{ref-105}, ColonDB \cite{ref-104}, and ETIS \cite{ref-102}. Note that we still utilize the Depth Anything Model to generate pseudo depth maps. We follow the training set setup of previous works \cite{ref-106, ref-103, ref-102}, and the experiment details are the same as those of our SATNet. We compare our SATNet with five SOTA methods, including PraNet \cite{ref-102}, CTNet \cite{ref-103}, SEPNet \cite{ref-106}, BUNet \cite{ref-104}, and CGLCON \cite{ref-105}. The comparison results are reported in Table \uppercase\expandafter{\romannumeral11}. Compared with the heavyweight segmentation methods, such as CGLCON \cite{ref-105} and PraNet \cite{ref-102}, our SATNet can achieve competitive performance with only 1.5G FLOPS. Besides, SATNet only uses the tiny Swin Transformer to obtain SOTA results. These results demonstrate that our SATNet model maintains the same advantages for polyp segmentation.
 
\subsubsection{RGB-Thermal SOD}
RGB-Thermal salient object detection (RGB-T SOD) aims to pinpoint prominent objects within aligned pairs of visible and thermal infrared images. To verify whether our SATNet can also achieve a balance between speed and accuracy on RGB-T data, we train and test our SATNet on RGB-T datasets, including VT821 \cite{ref-114}, VT1000 \cite{ref-113}, and VT5000 \cite{ref-111}. We follow the training set configuration of previous works, and the experiment details are the same as those of our SATNet. We compare our SATNet with five SOTA methods, including CGFNet \cite{ref-111}, ECFFNet \cite{ref-113}, SwinNet \cite{ref-35}, TriTransNet \cite{ref-115}, and XMSNet \cite{ref-116}. The comparison results are reported in Table \uppercase\expandafter{\romannumeral12}. Compared with these heavyweight methods, our SATNet can obtain competitive performance with fewer FLOPs, which proves that our SATNet still achieves the balance between speed and accuracy on RGB-T data.

 \section{\textbf{Conclusion}}
 In this paper, we propose a Speed-Accuracy Tradeoff Network, named SATNet. Firstly, we leverage the Depth Anything Model as a powerful zero-shot vision foundation model to generate high-quality depth maps, which avoids the adverse impact of original poor depth maps. Then, we design a Decoupled Attention Module (DAM) specifically for the lightweight paradigm, which decouples multi-modality features into horizontal and vertical vectors to learn cues of the two views via an alternated manner. Next, we propose a Dual Information Representation Module (DIRM) to represent texture and saliency features, which enriches the constrained feature space and enhance the representation ability of features in a lightweight framework. Finally, we devise a Dual Feature Aggregation Module (DFAM) to integrate texture and saliency features. Our method successfully improves the performance of lightweight RGB-D SOD and achieves the balance between efficiency and accuracy.

 With the emergence of vision foundation models, numerous tasks in computer vision have witnessed new opportunities and advancements. The Segment Anything Model, as the first foundation model for segmentation tasks, has significantly propelled the development of fundamental tasks such as semantic segmentation and instance segmentation. To align with the development of visual foundation models, introducing large models such as SAM into the field of Salient Object Detection (SOD) has emerged as a noteworthy direction. However, as we integrate SAM into this domain, it is crucial to pay close attention to the new challenges that it brings, e.g., over-segmentation and mask ambiguity. In future work, we will attempt to integrate large - scale models with the SOD community by leveraging techniques such as distillation or fine-tuning of vision foundation models, and abandon the current non-learning prompt strategy of SAM.
 

\ifCLASSOPTIONcaptionsoff
\newpage
\fi

\bibliographystyle{./IEEEtran}
\bibliography{./IEEEabrv,./IEEEexample}
\begin{IEEEbiography}[{\includegraphics[width=1in,height=1.25in,clip,keepaspectratio]{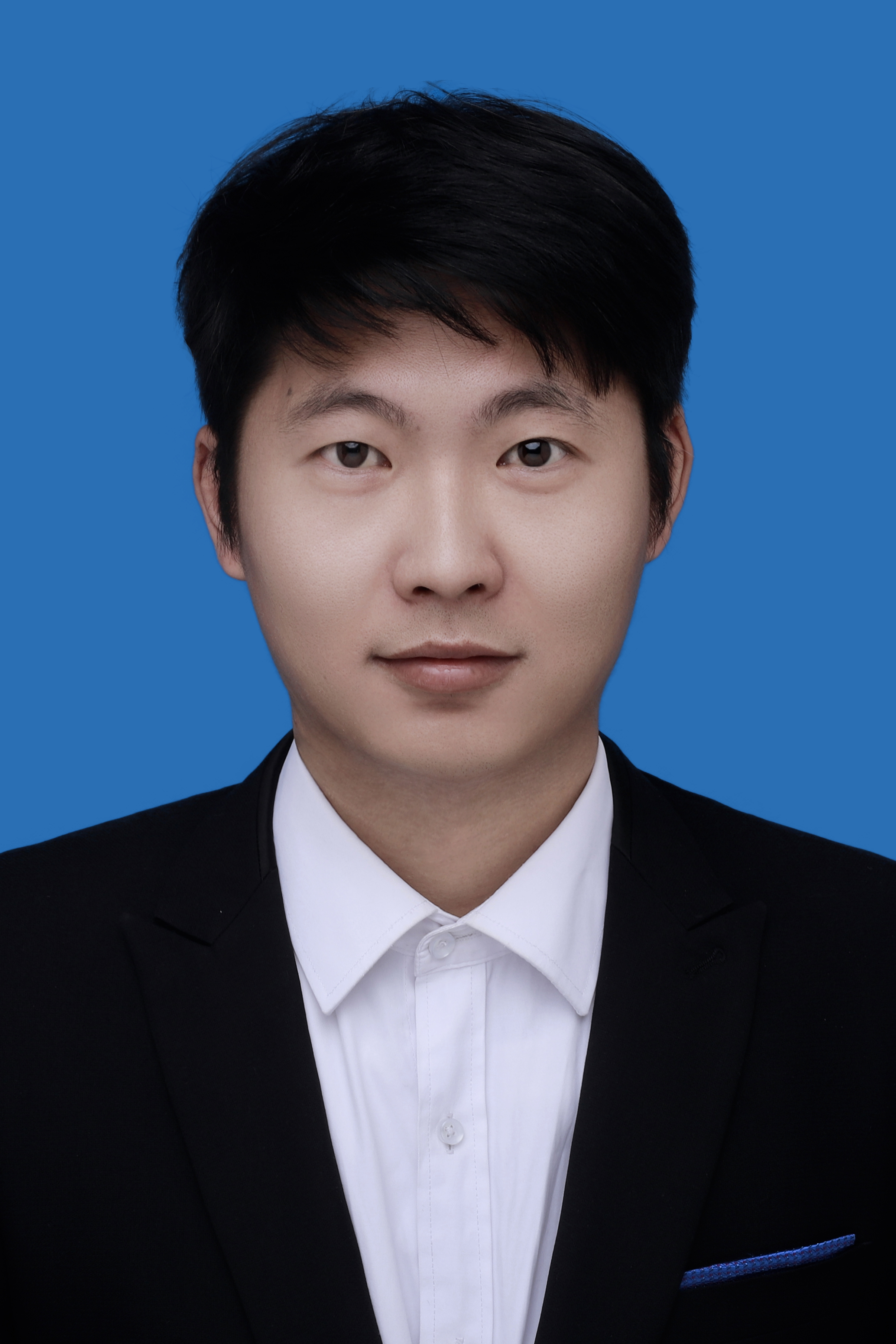}}]{Songsong Duan}
(Student Member, IEEE) received the M.S. degree with Anhui University of Science and Technology, Huainan, China in 2023. He is currently pursuing the Ph.D. degree with the School of Telecommunications Engineering, Xidian University, Xi’an, China. His research interests include computer vision, weakly supervised learning, and open-vocabulary learning.
\end{IEEEbiography}
\begin{IEEEbiography}[{\includegraphics[width=1in,height=1.25in,clip,keepaspectratio]{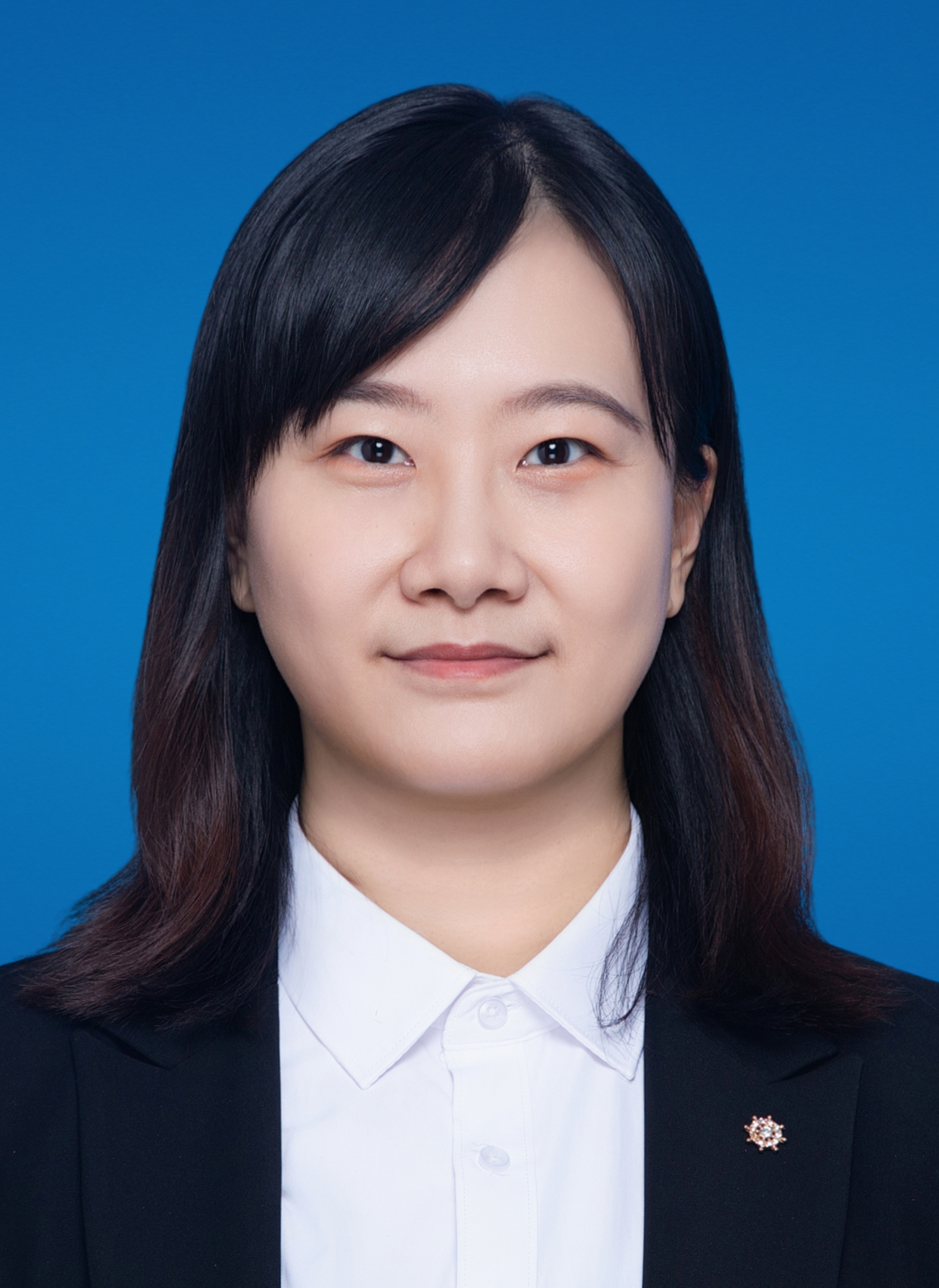}}]{Xi Yang}
(Senior Member, IEEE) received the B.Eng. degree in electronic information engineering and the Ph.D. degree in pattern recognition and intelligence system from Xidian University, Xi’an, China, in 2010 and 2015, respectively. From 2013 to 2014, she was a Visiting Ph.D. Student with the Department of Computer Science, The University of Texas at San Antonio, San Antonio, TX, USA. In 2015, she joined the State Key Laboratory of Integrated Services Networks, School of Telecommunications Engineering, Xidian University, where she is currently a Professor of communications and information systems. She has published over 60 articles in refereed journals and proceedings, including
IEEE TRANSACTIONS ON IMAGE PROCESSING, IEEE TRANSACTIONS ON NEURAL NETWORKS AND LEARNING SYSTEMS, IEEE TRANSACTIONS ON CYBERNETICS, IEEE TRANSACTIONS ON GEOSCIENCE AND REMOTE SENSING, CVPR, ICCV, and ACM MM. Her current research interests include image/video processing, computer vision, and machine learning.
\end{IEEEbiography}
\begin{IEEEbiography}[{\includegraphics[width=1in,height=1.25in,clip,keepaspectratio]{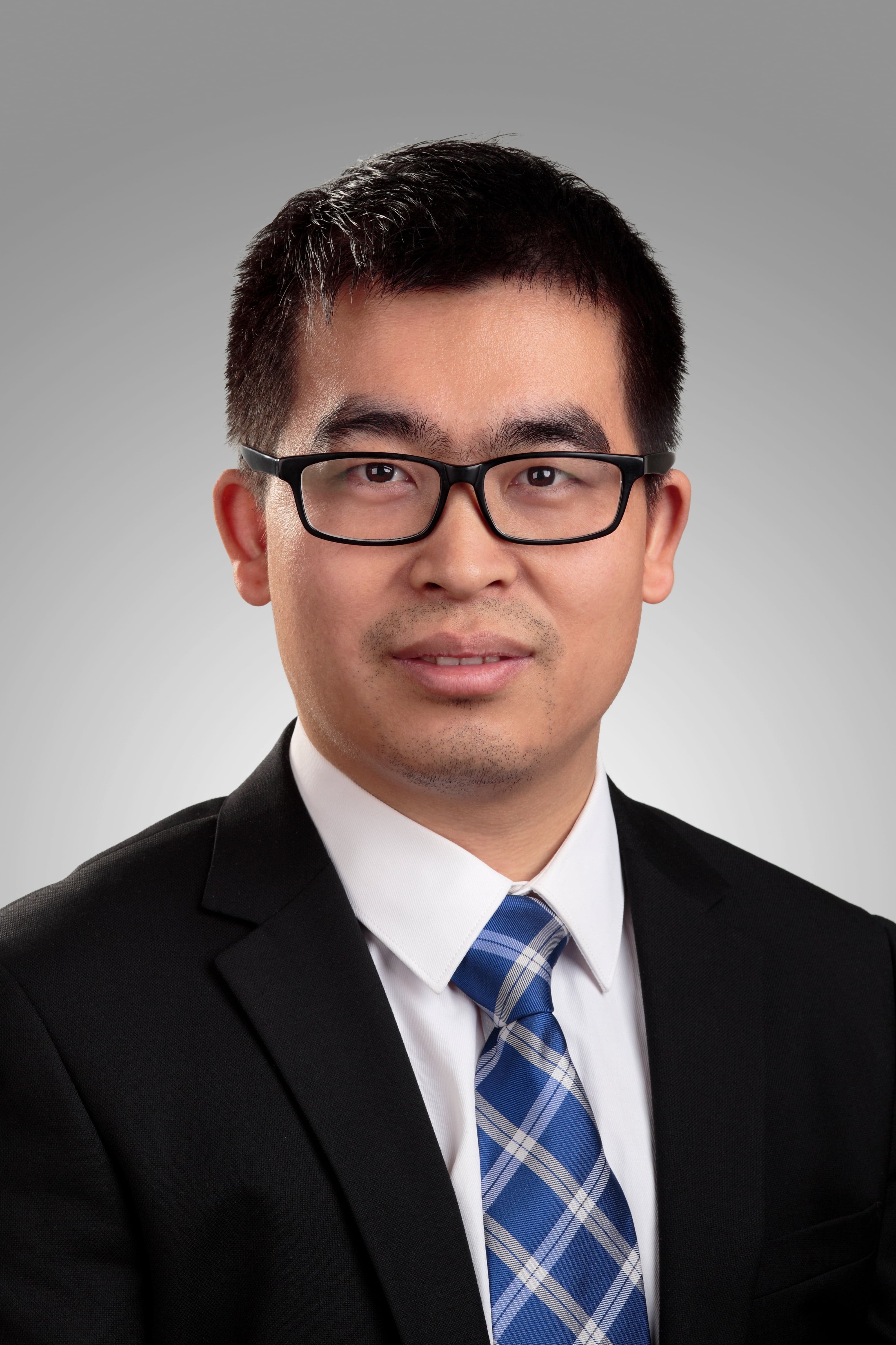}}]{Nannan Wang}
(Senior Member, IEEE) received the B.Sc. degree in information and computation science from Xi’an University of Posts and Telecommunications in 2009 and the Ph.D. degree in information and telecommunications engineering from Xidian University in 2015. He is currently a Professor with the State Key Laboratory of Integrated Services Networks, Xidian University. He has published over 150 articles in refereed journals and proceedings, including IEEE TRANSACTIONS ON PATTERN ANALYSIS AND MACHINE INTELLIGENCE, IJCV, CVPR, and ICCV. His current research interests include computer vision and machine learning.
\end{IEEEbiography}
\begin{IEEEbiography}[{\includegraphics[width=1in,height=1.25in,clip,keepaspectratio]{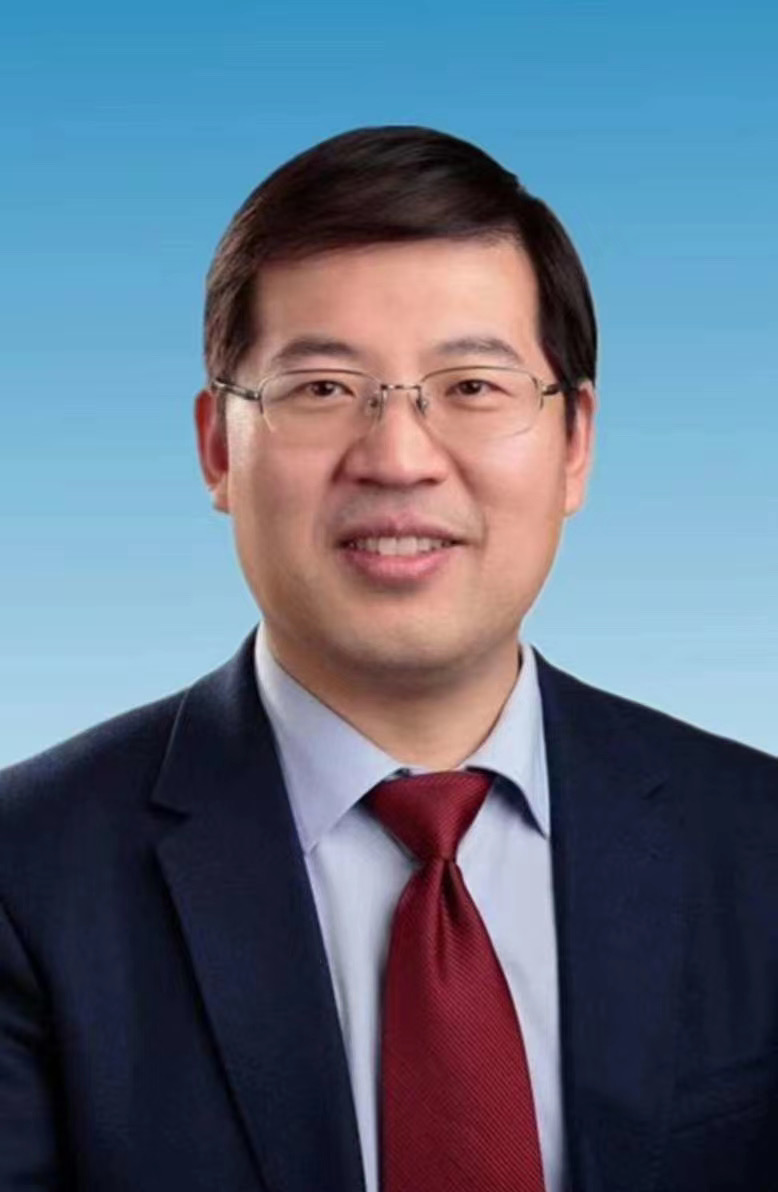}}]{Xinbo Gao}
(Fellow, IEEE) received the B.Eng., M.Sc., and Ph.D. degrees in electronic engineering, signal and information processing from Xidian University, Xi’an, China, in 1994, 1997, and 1999, respectively. From 1997 to 1998, he was a Research Fellow with the Department of Computer Science, Shizuoka University, Shizuoka, Japan. From 2000 to 2001, he was a Post-Doctoral Research Fellow with the Department of Information Engineering, The Chinese University of Hong Kong, Hong Kong. Since 1999, he has been with the School of Electronic Engineering, Xidian University, where he is currently a Professor of pattern recognition and intelligent system. Since 2020, he has been also a Professor of computer science and technology with Chongqing University of Posts and Telecommunications. He has published seven books and around 300 technical articles in refereed journals and proceedings. His current research interests include computer vision, machine learning, and pattern recognition. He is a fellow of IET, AAIA, CIE, CCF, and CAAI. He served as the general chair/co-chair, the program committee chair/co-chair, and a PC member for around 30 major international conferences. He is on the Editorial Boards of several journals, including Signal Processing (Elsevier) and Neurocomputing (Elsevier).
\end{IEEEbiography}
\end{document}